\documentclass[10pt,twocolumn,letterpaper]{article}

\usepackage[pagenumbers]{cvpr} 

\usepackage[dvipsnames]{xcolor}

\renewcommand{\arraystretch}{1}

\definecolor{cvprblue}{rgb}{0.21,0.49,0.74}
\usepackage[accsupp]{axessibility}
\usepackage[pagebackref,breaklinks,colorlinks,citecolor=cvprblue]{hyperref}
\usepackage{multicol}
\usepackage{multirow}
\usepackage{amsmath}
\usepackage{color}
\usepackage{subcaption}
\usepackage{bbding}
\usepackage{makecell}
\usepackage{xcolor,colortbl}
\usepackage[misc]{ifsym}

\def\paperID{3646} 
\def\confName{CVPR}
\def\confYear{2024}

\newcommand*{\affaddr}[1]{#1} 
\newcommand*{\affmark}[1][*]{\textsuperscript{#1}}

\makeatletter
\newcommand{\printfnsymbol}[1]{\textsuperscript{\@fnsymbol{#1}}}
\makeatother

\makeatletter
\def\thanks#1{\protected@xdef\@thanks{\@thanks
        \protect\footnotetext{#1}}}
\makeatother

\title{Exploring the Transferability of Visual Prompting for Multimodal \\Large Language Models\vspace{-1ex}}

\author{
Yichi Zhang\affmark[1,2]\quad
Yinpeng Dong\affmark[1,2]\textsuperscript{\Letter}\quad
Siyuan Zhang\affmark[1]\quad
Tianzan Min\affmark[1]\quad
Hang Su\affmark[1,3]\textsuperscript{\Letter}\quad
Jun Zhu\affmark[1,2,3]\thanks{\textsuperscript{\Letter}~Corresponding authors. Code available at \url{https://github.com/zycheiheihei/Transferable-Visual-Prompting}}\quad
\\\affaddr{\affmark[1]Dept. of Comp. Sci. and Tech., Institute for AI, Tsinghua-Bosch Joint ML Center,\\
THBI Lab, BNRist Center, Tsinghua University, Beijing 100084, China\\ \affmark[2]RealAI \quad \affmark[3] Pazhou Laboratory (Huangpu), Guangzhou, Guangdong 
}
\\
\tt\small{\{zyc22@mails., dongyinpeng@, siyuan-z20@mails., suhangss@, dcszj@\}tsinghua.edu.cn}
\vspace{-4ex}
  }

\begin{document}
\maketitle
\begin{abstract}
Although Multimodal Large Language Models (MLLMs) have demonstrated promising versatile capabilities, their performance is still inferior to specialized models on downstream tasks, which makes adaptation necessary to enhance their utility. However, fine-tuning methods require independent training for every model, leading to huge computation and memory overheads. In this paper, we propose a novel setting where we aim to improve the performance of diverse MLLMs with a group of shared parameters optimized for a downstream task. To achieve this, we propose Transferable Visual Prompting (TVP), a simple and effective approach to generate visual prompts that can transfer to different models and improve their performance on downstream tasks after trained on only one model. We introduce two strategies to address the issue of cross-model feature corruption of existing visual prompting methods and enhance the transferability of the learned prompts, including 1) Feature Consistency Alignment: which imposes constraints to the prompted feature changes to maintain task-agnostic knowledge; 2) Task Semantics Enrichment: which encourages the prompted images to contain richer task-specific semantics with language guidance. We validate the effectiveness of TVP through extensive experiments with 6 modern MLLMs on a wide variety of tasks ranging from object recognition and counting to multimodal reasoning and hallucination correction.
\vspace{-3ex}
\end{abstract}
\section{Introduction}
\label{sec:intro}

\begin{figure}[t]
    \centering
    \includegraphics[width=\linewidth]{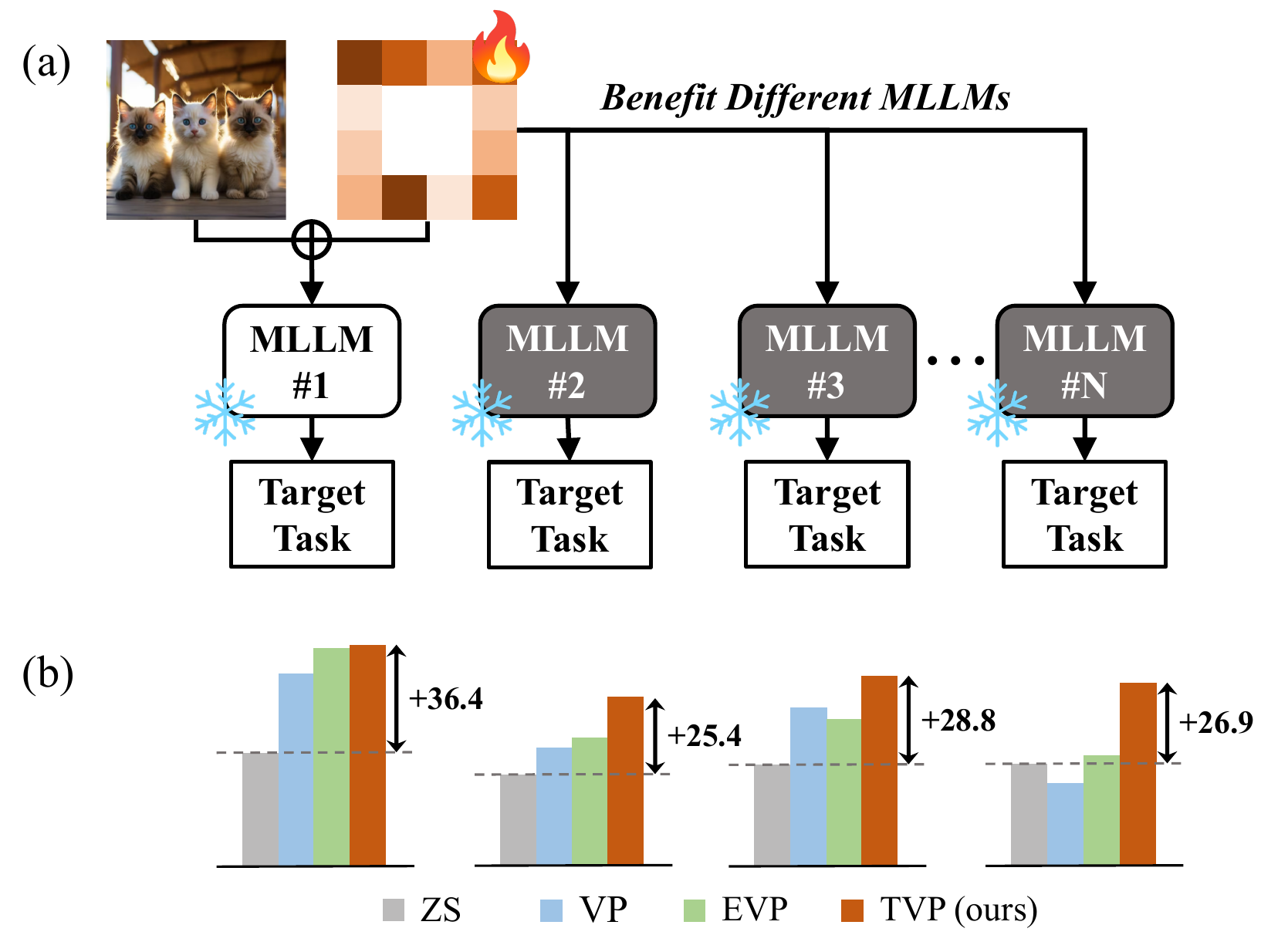}
    \vspace{-4ex}
    \caption{\textbf{(a) Illustration of problem setting:} We aim to improve the performance  of different MLLMs on a specific task with a set of shared parameters. This is achieved by exploiting the transferability of the visual prompts trained on one model and using them on other models. 
    \textbf{(b) Demonstration of the effect:} We show the partial results on SVHN~\cite{svhn} with the visual prompt trained on MiniGPT-4~\cite{zhu2023minigpt} and tested on InstructBLIP~\cite{Dai2023InstructBLIP}, BLIP2~\cite{li2023blip} and BLIVA~\cite{hu2023bliva}. Compared with the existing visual prompting methods~\cite{bahng2022exploring,wu2022unleashing}, the proposed Transferable Visual Prompting (TVP) improves different models with larger margins. Detailed results are in~\cref{sec:main_results}. ZS is for zero-shot inference when non-prompted.}
    \vspace{-2ex}
    \label{fig:first}
\end{figure}







The recent success of Large Language Models (LLMs)~\cite{touvron2023llama,roumeliotis2023chatgpt,touvron2023llama2} has motivated researchers to explore their capabilities in solving multimodal tasks. Tremendous efforts have been made to develop Multimodal Large Language Models (MLLMs)~\cite{zhu2023minigpt, li2023blip, Dai2023InstructBLIP,alayrac2022flamingo,liu2023visual}, which seamlessly integrate visual input into LLMs by aligning image features with text embeddings. 
These models have achieved remarkable performance in image understanding and reasoning \cite{fu2023mme, xu2023lvlm,li2023seed} and serve as ``foundation models'' \cite{bommasani2021opportunities} for a variety of tasks. 
Despite their excellent generalization performance, existing MLLMs usually lag behind the specialized state-of-the-art models on downstream tasks (\eg, image classification), especially when evaluated in zero-shot manner \cite{xu2023lvlm, zhai2023investigating}. This is because MLLMs are primarily pre-trained on massive data and fine-tuned on a small amount of modality alignment and instruction data \cite{liu2023visual,zhu2023minigpt, Dai2023InstructBLIP}, while lacking specialized training on certain tasks. Consequently, when users aim to employ MLLMs for downstream tasks, their performance is far from satisfactory, making it necessary to develop effective and efficient strategies to bridge this gap and enhance the utility of MLLMs in task-specific applications.


Adapting MLLMs for downstream tasks conventionally requires fine-tuning on task-specific data. Though effective in various fields, such as science~\cite{liu2023visual} and biomedicine \cite{li2023llava}, full-parameter fine-tuning (FFT) is computationally demanding and storage-intensive, particularly for models with billions of parameters. To address these problems, various parameter-efficient fine-tuning (PEFT) techniques have been proposed, including Adapters~\cite{houlsby2019parameter}, LoRA~\cite{hu2021lora}, and prompt tuning \cite{jia2022vpt,liu2022p}. These methods perform gradient-based optimization of additional model-specific parameters for a downstream task. Nevertheless, they require a prohibitive amount of memory for optimization and the resultant parameters lack generalizability across different models.
In a practical scenario,  users with no prior knowledge of PEFT and limited computation resources would prefer auxiliary parameters (\eg, prompts) that can improve their own models on downstream tasks without further fine-tuning. This leads to a novel and challenging setting that we aim to develop a set of \emph{shared parameters} that can benefit numerous MLLMs on the same task, while only optimized on one or few of them, as shown in~\cref{fig:first}. This pathway is hopeful to be resource-friendly and flexible, making it easier to adapt different models for a given task simultaneously, even when model weights are not accessible. 
Moreover, it aligns with the ``Prompt as a Service'' (PaaS) paradigm \cite{wu2023quantifying}, where users can request a prompt for a downstream task from the PaaS provider while keeping the local
models confidential.

As MLLMs take images as input, the image pixel space is a promising shared space for parameter learning. Previous methods have explored visual prompting (VP)~\cite{bahng2022exploring,wu2022unleashing} to adapt pre-trained models for downstream tasks. VP learns parameters in pixel space around clean images as a frame, known as visual prompts.
Inspired by the transferability of adversarial examples \cite{dong2019evading,zhou2018transferable}, we take transferring visual prompts to boost the performance of other models as a feasible solution for our problem.
However, the transferability of the learned visual prompts for other models is yet to be studied. 
We find that, though VP can effectively elevate the performance of models used for prompt training, it can lead to limited performance improvement or significant degradation for other models.
We attribute this to the fact that the trained prompts lead to notable changes in visual features across different models, defined as \textit{cross-model feature corruption}. This indicates that the visual prompts overfit the model for their training and invalidate the plenty knowledge acquired from large-scale pre-training when transferred to other models, thus impacting their performance.


In this paper, we propose \textbf{Transferable Visual Prompting (TVP)} to enhance the transferability of visual prompts across MLLMs and improve these models simultaneously.
To achieve this, we formulate a unified framework of VP on different tasks for MLLMs. We propose two key strategies to fortify both general knowledge and task-specific representations. First, we propose \textbf{Feature Consistency Alignment (FCA)} to mitigate the issue of feature corruption that highly suppresses the transferability. FCA facilitates model adaptation to downstream tasks by imposing constraints on visual features after applying prompts, preserving essential inner knowledge.
Consequently, it helps models better retain and leverage task-agnostic representations for improvement.
Second, we introduce \textbf{Task Semantics Enrichment (TSE)} to further embed task information explicitly into visual prompts. 
By leveraging CLIP~\cite{radford2021clip}, TSE encourages the prompted images to exhibit semantic similarity with text features tailored for specific tasks, rather than simply utilizing task-specific objectives for end-to-end prompt learning. This enables models to extract better shareable task-specific semantics and get improved on the target tasks.


We examine the performance of TVP through substantial experiments. The visual prompts trained with one single model can facilitate the overall performance of 6 modern MLLMs on 10 datasets ranging from visual tasks like recognition and counting to multimodal reasoning and hallucination correction, which significantly surpasses the existing visual prompting baselines. The performance is further improved with model ensembling. We demonstrate that TVP can enhance different models with diverse data scales, generalize to different datasets, and resist image corruptions, 
emphasizing the practicality of our method in real scenarios. Comparisons with existing fine-tuning methods suggest the feasibility of improving different models with shared parameters and the effectiveness of our TVP.


\begin{figure*}[t]
    \centering
    \includegraphics[width=\linewidth]{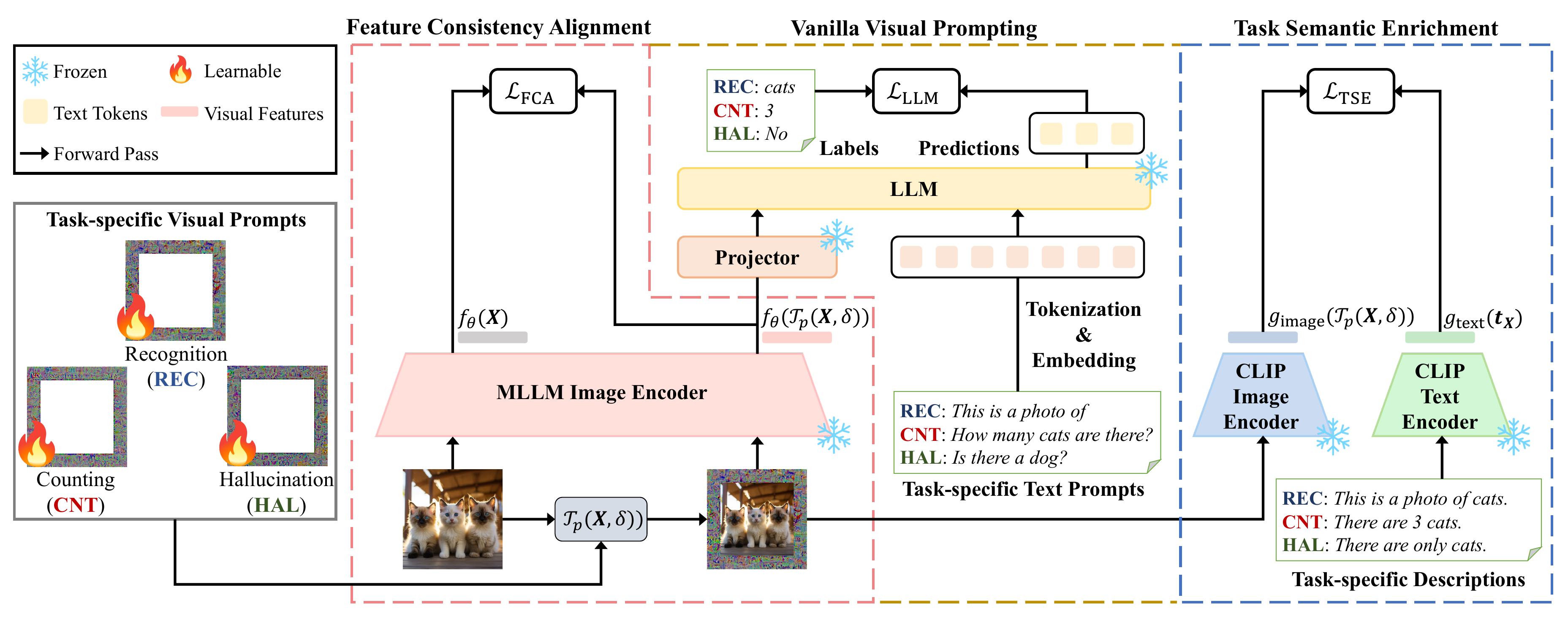}
    \vspace{-5ex}
    \caption{Overview of our proposed Transferable Visual Prompting (TVP) method for adapting MLLMs. TVP optimizes a visual prompt on a single MLLM towards a downstream task. Feature Consistency Alignment (FCA) and Task Semantic Enrichment (TSE) are proposed to make learned visual prompts more transferable and benefit more unseen MLLMs to improve on the same task.}
    \vspace{-2ex}
    \label{fig:pipeline}
\end{figure*}

\section{Related Works}

In this section, we briefly review the related works in the context of Multimodal Large Language Models and adaptation methods for large-scale pre-trained models.

\subsection{Multimodal Large Language Models}

The significant advancements in LLMs for language-centric tasks have spurred investigations into their potential applications in diverse multimodal contexts~\cite{yin2023survey}. This exploitation is primarily manifested in works focusing on modality alignment and instruction tuning~\cite{alayrac2022flamingo,li2023blip,liu2023visual,zhu2023minigpt,Dai2023InstructBLIP}. These proposed models lay the foundation for MLLMs and many subsequent works have been proposed to improve performance concerning the issues of in-context learning~\cite{zhao2023mmicl}, efficient training~\cite{zhang2023transfer}, richer modalities~\cite{su2023pandagpt}, \etc.

Several benchmarks~\cite{li2023seed,xu2023lvlm, fu2023mme} have demonstrated that MLLMs show versatile capabilities in visual perception and comprehension. However, their performance falls short of specialized models on specific tasks, limiting their applicability in certain scenarios~\cite{xu2023lvlm}. Moreover, MLLMs face challenges related to safety and reliability, including issues of value alignment~\cite{liu2023trustworthy} and hallucination~\cite{li2023evaluating,gunjal2023detecting}. MLLMs need further tuning to address these challenges.

\subsection{Adaptation for Large-Scale Pre-trained Models}

Adapting MLLMs mainly follows methods for large models (\eg, LLMs~\cite{touvron2023llama}, CLIP~\cite{radford2021clip}). Fine-tuning for a downstream task is straightforward but costly in computation and storage. Parameter-efficient fine-tuning (PEFT) methods, such as Adapters~\cite{houlsby2019parameter}, LoRA~\cite{hu2021lora}, and prompt tuning \cite{jia2022vpt,liu2022p}, have emerged to ease these challenges. Some recent advanced works also focus on efficient modality bridging and adaptation via routing and skipping with adapters~\cite{luo2023cheap,wu2023parameter}. However, they are inherently model-specific and require access to the inner structure of models, diverging from our goal of optimizing a single set of parameters to adapt multiple models in a resource-friendly and flexible manner.

Recent developments in visual prompting~\cite{bahng2022exploring}, inspired by adversarial reprogramming~\cite{elsayed2018adversarial,tsai2020transfer}, offer a promising solution for model adaptation by introducing learnable perturbations in the pixel space of images. 
As the pixel space is a shared domain for different models, it becomes a natural choice for parameter tuning. 
Many follow-up works have explored topics like performance refinement~\cite{wu2022unleashing} and data generalization~\cite{huang2023diversity,khattak2023self}, but none has studied the generalization of visual prompts across models, or their transferability as defined in adversarial attacks~\cite{dong2018boosting,zhou2018transferable}. 
While popular works of prompt tuning like CoOp~\cite{zhou2022learning,zhou2022conditional}, VPT~\cite{jia2022vpt} and MaPLe~\cite{khattak2023maple} operate soft prompts for both modalities at the early layers of the model, even at the embedding space, they are invalid under complete black-box conditions where only discrete texts and images are accessible for input.

In this paper, we investigate the direct transfer of trained visual prompts to other MLLMs for adaptation. This reduces the computation and storage overloads, and also offers a more convenient and flexible solution in diverse application scenarios like ``Prompt as a Service'' (PaaS)~\cite{wu2023quantifying}, where users can directly request a visual prompt towards a certain task for their local models from the PaaS provider with a guarantee of the model confidentiality.


\section{Methods}

Visual prompting offers an effective means to adapt vision-language models, such as CLIP~\cite{radford2021clip}, to downstream visual tasks without resorting to fine-tuning. In this study, we extend the application of VP to MLLMs and investigate its potential for enhancing performance across a range of models. Although existing methods can enhance model performance through prompt training, these trained prompts often fall short when applied to other models due to issues related to feature corruption. To this end, we introduce the method of Transferable Visual Prompting (TVP), aiming to enhance the transferability of visual prompts across diverse MLLMs.

In this section, we will first briefly present some preliminaries about MLLMs and VP, then formulate our problem of transferring visual prompts across MLLMs, and finally introduce our proposed TVP approach. The overview of our method is depicted in~\cref{fig:pipeline}.

\subsection{Preliminaries}
\label{sec:prelim}
\textbf{Multimodal Large Language Models.} MLLMs primarily use an architecture that projects visual features to the text embedding space to integrate images with LLMs~\cite{zhu2023minigpt,Dai2023InstructBLIP,hu2023bliva}. 

To be specific, assume that we have a visual encoder $f_\theta$, an LLM $P_\phi$ and a projector $h_\psi$. The textual response $\mathbf{r}$ of an MLLM given image input $\mathbf{X} $ and text input $\mathbf{t}$ is decided autoregressively according to the likelihood
\begin{equation}
   \mathbf{r}_i\sim P_\phi(\mathbf{r}_i|h_\psi(f_\theta(\mathbf{X})),\mathbf{t}, \mathbf{r}_{<i}),
\end{equation}
where $\mathbf{X}\in\mathbb{R}^{3\times H\times W}$ is an RGB image and $\mathbf{t}\in \mathbb{V}^{N}$ is a text with $N$ tokens from vocabulary $\mathbb{V}$. Image features $f_\theta(\mathbf{X})$ are mapped by a projector $h_\psi$ (\eg, MLP~\cite{zhu2023minigpt,Dai2023InstructBLIP,liu2023visual}) to align with the text and further concatenated with text tokens as unified input for the downstream LLM.

\textbf{Visual Prompting.} As proposed in~\cite{elsayed2018adversarial} and~\cite{bahng2022exploring}, a trainable visual prompt $\boldsymbol{\delta}\in \mathbb{R}^{3\times H\times W}$ is learned in the pixel space and imposed to the clean images with different transformations $\mathcal{T}$ (\eg, global perturbations~\cite{oh2023blackvip}, padding~\cite{bahng2022exploring}) to adapt models to a certain downstream task. 


In this paper, we follow the common practice of VP~\cite{bahng2022exploring} by adding universal pixel-level prompt around resized input images. Mathematically, we describe the process of visual prompting as
\begin{equation}
    \mathcal{T}_p(\mathbf{X}, \boldsymbol{\delta}) = \text{Resize}_{H\times W\rightarrow H'\times W'}(\mathbf{X}) +\underbrace{M_p \odot \boldsymbol{\delta}}_{\text{visual prompt}},
\end{equation}
where $p$ is the width of the visual prompt and $M_p$ is a binary mask with a border of width $p$ taking values of $1$. The original image $\mathbf{X}$ of size $H\times W$ is resized to $H'\times W'=(H-2p)\times(W-2p)$, so that the prompted image is of the same size as the original without overlapping with $\boldsymbol{\delta}$. We take $H=W=224$ and $p=30$ by default.

\subsection{Problem Formulation}
\label{sec:vp4mllm}

We extend VP~\cite{bahng2022exploring,wu2022unleashing} to adapt MLLMs for downstream tasks while avoiding heavy computations in massive parameter fine-tuning. 
To make it more general, we unify different visual tasks into the form of text completion and take the autoregressive loss (\ie, cross-entropy loss over the vocabulary) as the training objective for VP. For a task with dataset $\mathcal{D}$, this is formulated as minimizing $\mathcal{L}_{\text{LLM}}(\boldsymbol{\delta})=$
\begin{align}
     \underset{(\mathbf{X},\mathbf{t},\mathbf{r})\sim\mathcal{D}}{\mathbb{E}}\left[\sum_{i=1}^{N_r}-\log P_\phi(\mathbf{r}_i|h_\psi(f_\theta(\mathcal{T}_p(\mathbf{X},\boldsymbol{\delta}))),\mathbf{t}, \mathbf{r}_{<i})\right].
\label{eq:formulation}
\end{align}
Here, $(\mathbf{t}, \mathbf{r})$ is the prompt-target text pair for a task, with $N_r$ denoting the length of $\mathbf{r}$. Prompts and targets for different tasks are introduced in~\cref{sec:datasets}.

In this work, we exploit the transferability of visual prompts, inspired by transfer attacks in the field of adversarial robustness~\cite{zhou2018transferable}. We transfer the one-time trained prompts to other models to improve their performance. Specifically, we optimize a prompt $\boldsymbol{\delta}$ on an MLLM minimizing its loss $\mathcal{L}_{\text{LLM}}$ and expect it can lower the loss $\mathcal{L}'_{\text{LLM}}$ of an arbitrary different MLLM, \ie, to improve its performance when we apply this trained prompt on it without any further fine-tuning on the target task.


\begin{figure}[t]
    \centering
    \begin{subfigure}{.5\linewidth}
        \centering
        \includegraphics[width=.9\linewidth]{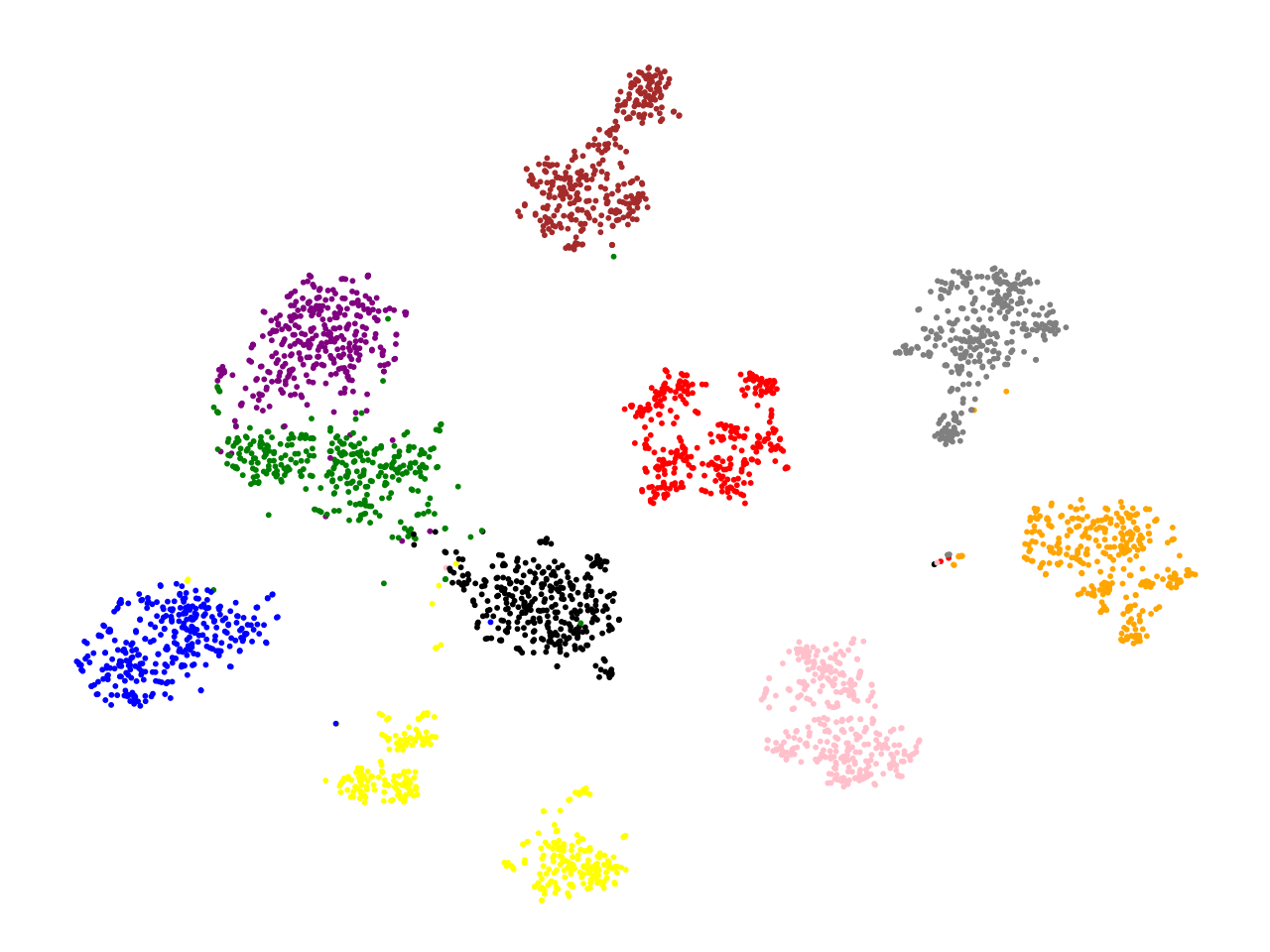}
        \caption{InstructBLIP-ZS-88.11\%}
        \label{fig:sub1}
    \end{subfigure}%
    \begin{subfigure}{.5\linewidth}
        \centering
        \includegraphics[width=.9\linewidth]{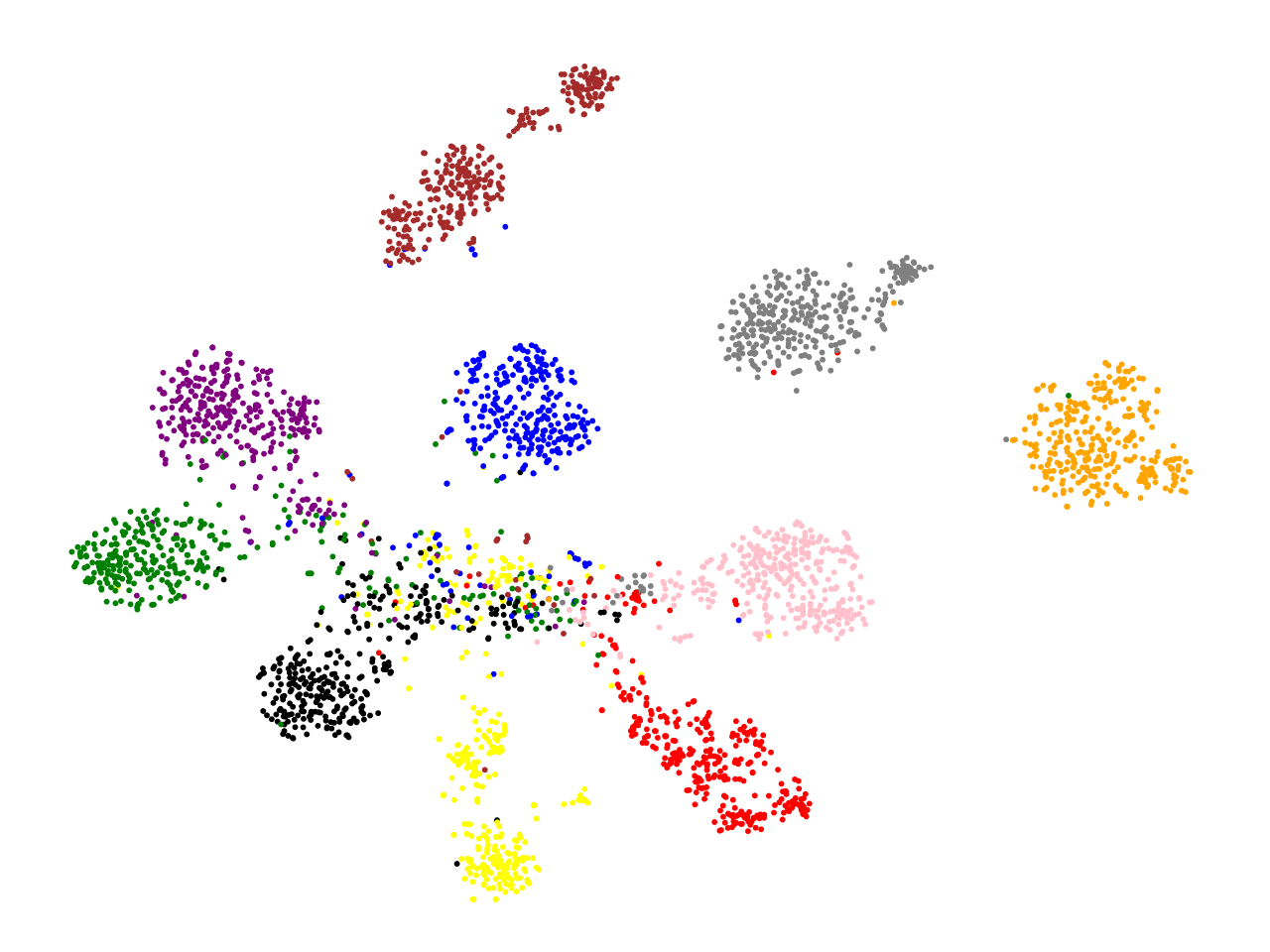}
        \caption{InstructBLIP-Prompted-82.07\%}
        \label{fig:sub2}
    \end{subfigure}
    \newline
    \begin{subfigure}{.5\linewidth}
        \centering
        \includegraphics[width=.9\linewidth]{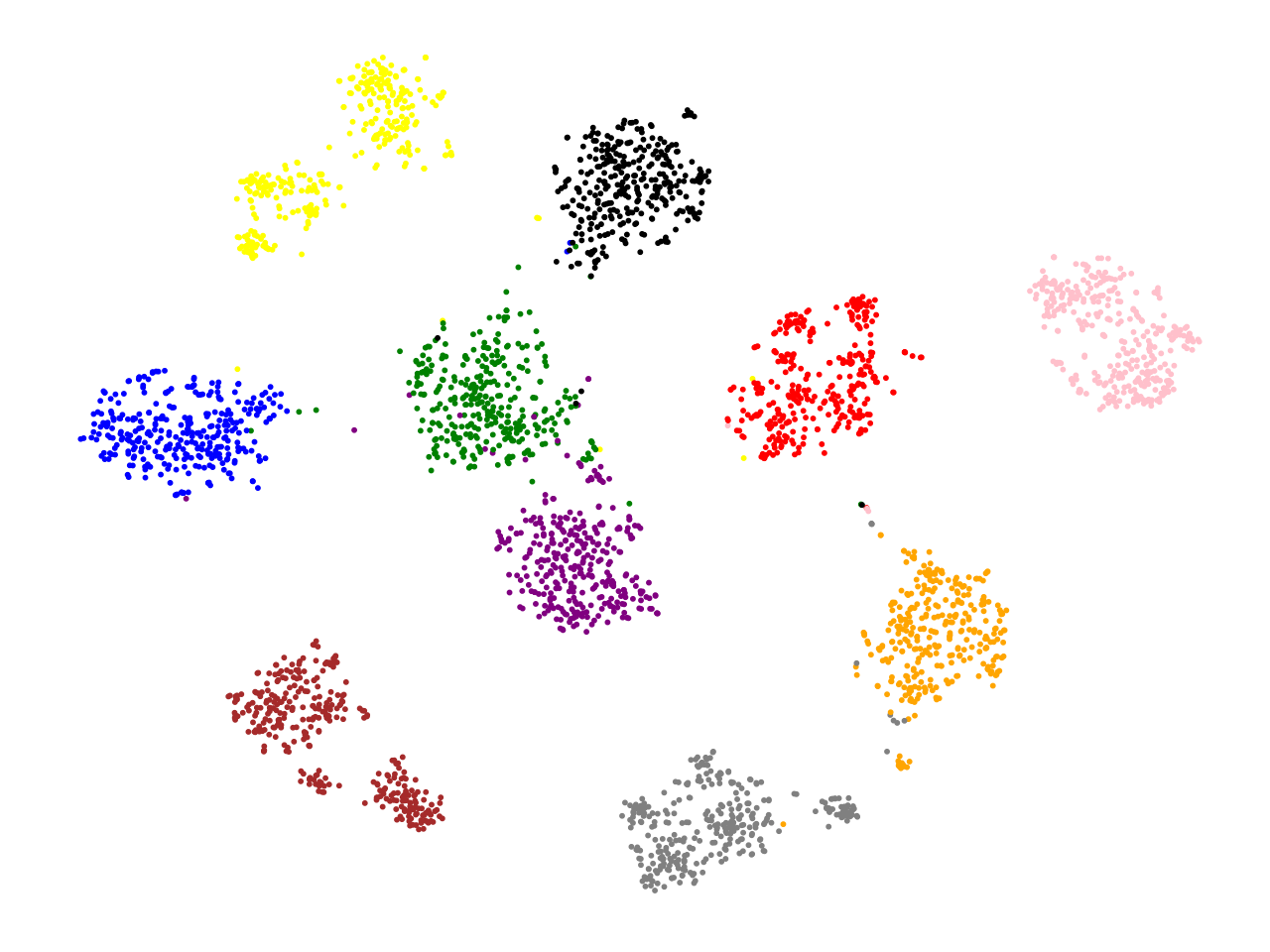}
        \caption{BLIVA-ZS-89.23\%}
        \label{fig:sub3}
    \end{subfigure}%
    \begin{subfigure}{.5\linewidth}
        \centering
        \includegraphics[width=.9\linewidth]{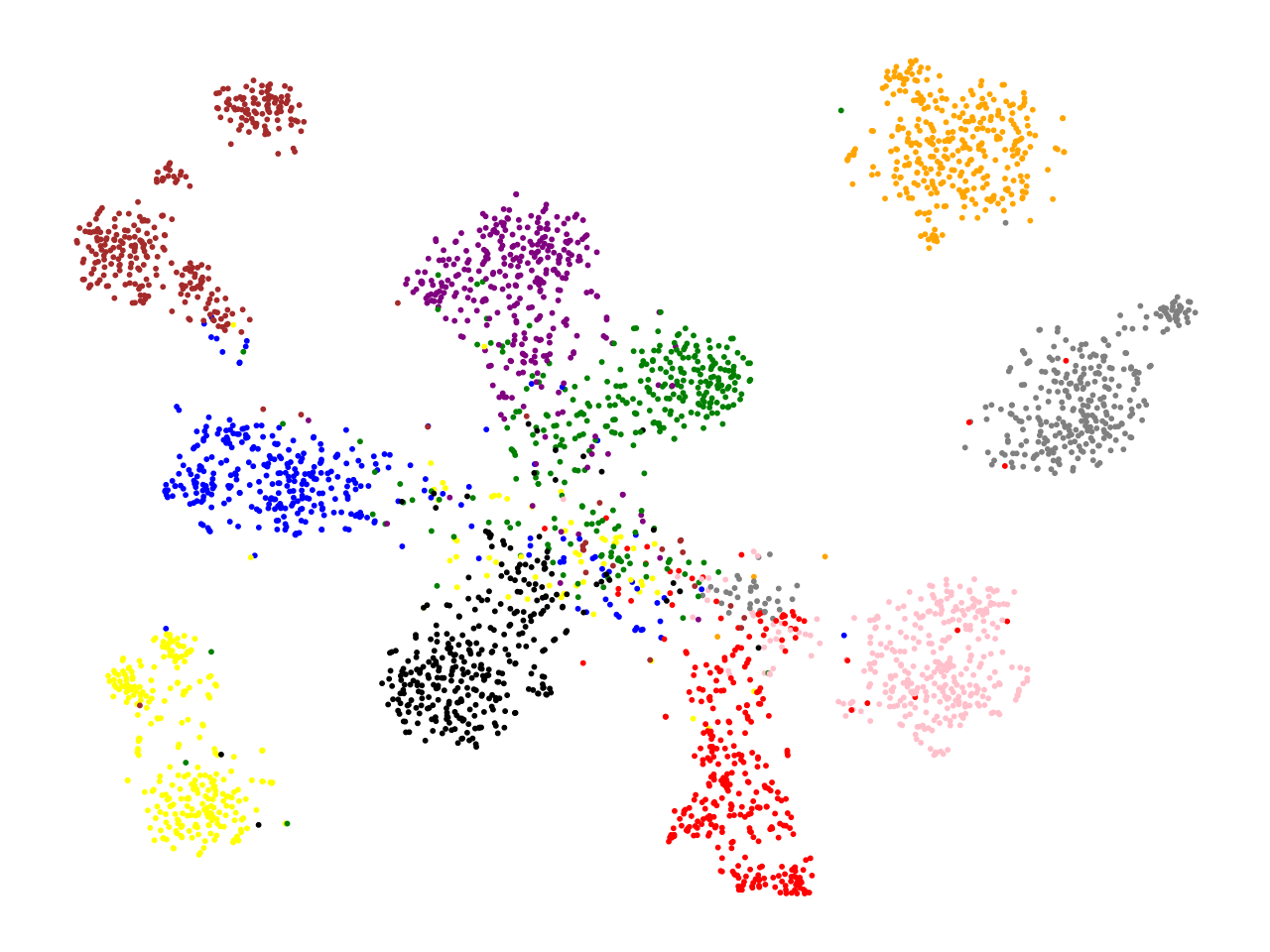}
        \caption{BLIVA-Prompted-78.85\%}
        \label{fig:sub4}
    \end{subfigure}
    \caption{t-SNE visualization of visual features from InstructBLIP and BLIVA on CIFAR-10 with and without the visual prompt, which is trained on MiniGPT-4 using VP~\cite{bahng2022exploring}. When the images are prompted, the visual features of different categories get mixed together, leading to performance degradation.}
    \vspace{-3ex}
    \label{fig:test}
\end{figure}

After examining existing VP methods~\cite{bahng2022exploring,wu2022unleashing} on different models, we find that the transferability of the generated visual prompts is poor, resulting in modest improvement or even remarkable performance decline, as shown in~\cref{fig:first}. We identify the reason for this phenomenon as ``\textit{cross-model feature corruption}'' given visual prompts trained on one model.
We can observe significant changes in visual features triggered on different models by the visual prompts. As shown in \cref{fig:test}, by plotting the t-SNE~\cite{van2008visualizing} of prompted visual features extracted by different models on CIFAR-10, we find that for models with performance degradation, the features of prompted images get mixed up compared to clean images. It indicates that when training visual prompts, they primarily amplify task-specific features that are only useful for the current model, \ie, overfitting to the model for training. However, the feature changes on other models render the knowledge from pre-training ineffective and disrupt the predictions for those models without prompt learning.

\subsection{Transferable Visual Prompting}

To alleviate the issue of feature corruption and further improve the transferability, we present Transferable Visual Prompting (TVP) by integrating two novel strategies with traditional VP techniques. We introduce them as follows.

\subsubsection{Feature Consistency Alignment}

The above analysis reveals that visual prompts significantly alter image features, leading to a loss of the general knowledge gained from pre-training and a reduction of performance when transferring across models.
To counter this, we propose to impose a specific constraint on the divergence between the prompted features and non-prompted features. This is intended to avoid exceptional feature corruption and guide the prompted features to maintain the task-agnostic general knowledge during prompt learning. 

We encourage the prompted features to be consistent with original features by a loss of feature consistency alignment (FCA), so that the task-agnostic features and inherent knowledge can be aligned. Given a white-box model $(P_\phi, f_\theta, h_\psi)$ and an input tuple $(\mathbf{X}, \mathbf{t}, \mathbf{r})$, we will get a feature of the plain image from the visual encoder $f_\theta(\mathbf{X})$ and a prompted feature $f_\theta(\mathcal{T}_p(\mathbf{X},\boldsymbol{\delta}))$ accordingly. The $\ell_2$ distance between these features are adopted to measure the divergence and the FCA loss can be computed as $\mathcal{L}_{\text{FCA}}(\boldsymbol{\delta}) = $
\begin{equation}
    \underset{(\mathbf{X},\mathbf{t},\mathbf{r})\sim\mathcal{D}}{\mathbb{E}} \left[\frac{1}{2}\|f_\theta(\mathcal{T}_p(\mathbf{X},\boldsymbol{\delta}))-f_\theta(\mathbf{X})\|^2_2\right].
\end{equation}
Here, we condition the prompted features with original features to depress changes in features and preserve effective task-agnostic semantic information. It is expected to make the learned visual prompts more transferable because the prompts are more mild to exploit useful visual representations from unseen models. 

From another perspective, the FCA loss can be considered as a form of regularization applied to the prompted feature space, serving as a means of enhancing generalization. Regularization techniques have been commonly used to prevent overfitting and guarantee the model's generalization to unseen data. The main difference is that previous works~\cite{khattak2023self,kim2021selfreg,van2017l2,zhou2022domain} primarily focus on the generalization across data (\eg, base-to-novel, domain generalization), while our research is centered around the generalization across models. By employing regularization to the prompted features, we seek a balance between maximizing the performance on supervised tasks and maintaining the inherent knowledge embedded in various models.

\subsubsection{Task Semantic Enrichment}

Besides end-to-end supervised training, we want to explicitly make visual prompts contain richer task-specific semantic information to further boost the performance of visual prompts. This is expected to enhance the performance of diverse models on the target tasks by fostering a shared semantic enhancement across them.
CLIP~\cite{radford2021clip} is a vision-language foundation model and has abundant knowledge by image-text alignment. Based on this, many studies have taken CLIP as a bridge to introduce language as additional supervision and guidance for visual tasks~\cite{khandelwal2022simple,shen2021much}. Yet, in the context of prompt learning, it has not been widely studied to use CLIP as a guidance rather than the target model.

We propose another loss of task semantic enrichment (TSE) to explicitly enhance the task-related semantics of prompted images by leveraging CLIP. CLIP consists of a visual encoder $g_\text{image}$ and a text encoder $g_\text{text}$, mapping image input $\mathbf{X}\in\mathbb{R}^{3\times H\times W} $ and text input $\mathbf{t}\in \mathbb{V}^{N}$ to a shared embedding space $\mathbb{R}^d$ respectively. The correspondence between images and texts can be obtained by computing the distance between their features. By designing task-specific descriptions according to the images, we can maximize their similarity to better embed the task semantics into the prompted images. Referring to the contrastive loss of CLIP, we present an auxiliary loss as $\mathcal{L}_\text{TSE}(\boldsymbol{\delta}) =$
\begin{equation}
     \underset{(\mathbf{X},\mathbf{t}_{\mathbf{X}})\sim\mathcal{D}}{\mathbb{E}}[\exp(\tau\text{sim}(g_\text{image}(\mathcal{T}_p(\mathbf{X},\boldsymbol{\delta})),g_\text{text}(\textbf{t}_\mathbf{X})))],
\end{equation}
where $\text{sim}(\cdot,\cdot)$ is the cosine similarity and $\tau$ is the temperature. $\mathbf{t}_\mathbf{X}$ is the text description of image $\mathbf{X}$ under the target task. Descriptions for different tasks are in~\cref{sec:datasets}.


By integrating FCA loss and TSE loss along with supervised loss of $\mathcal{L}_\text{LLM}$ in~\cref{eq:formulation}, we guide the visual prompt to consolidate and strengthen task-agnostic and task-specific representations while improving the model performance. The overall training objective is in the form of
\begin{equation}
    \mathcal{L}(\boldsymbol{\delta}) = \mathcal{L}_\text{LLM}(\boldsymbol{\delta}) + \lambda_1 \mathcal{L}_\text{FCA}(\boldsymbol{\delta}) - \lambda_2 \mathcal{L}_\text{TSE}(\boldsymbol{\delta}),
\label{eq:final}
\end{equation}
where $\lambda_1$ and $\lambda_2$ are hyperparameters. For this training objective, we follow EVP~\cite{wu2022unleashing}, which introduces the concepts of input diversity and gradient normalization to improve the performance, and update the learnable prompt at step $t$ by
\begin{equation}
    \boldsymbol{\delta}^{t+1} = \boldsymbol{\delta}^t-\gamma \frac{\nabla_{\boldsymbol{\delta}^t}\mathcal{L}(\boldsymbol{\delta}^t)}{||\nabla_{\boldsymbol{\delta}^t}\mathcal{L}(\boldsymbol{\delta}^t)||_2},
\label{eq:update}
\end{equation}
where $\gamma$ is the learning rate. 

\section{Experiments}

In this section, we conduct substantial experiments to verify the effectiveness of the proposed TVP in boosting the transferability of visual prompts.

\begin{table*}  
    \centering
    \begin{subtable}{.495\linewidth}  
        \centering
        \scriptsize
        \setlength{\tabcolsep}{3pt}
        \resizebox{\linewidth}{!}{
        \begin{tabular}{c|c|cccccc>{\columncolor[gray]{0.9}}c}
        \toprule\toprule
        \multicolumn{2}{c|}{\bf Recognition: CIFAR-10} & \parbox[t]{2mm}{\rotatebox[origin=c]{90}{ MiniGPT-4}}  & \parbox[t]{2mm}{\rotatebox[origin=c]{90}{ InstructBLIP}}  & \parbox[t]{2mm}{\rotatebox[origin=c]{90}{ BLIP2}}  & \parbox[t]{2mm}{\rotatebox[origin=c]{90}{ VPGTrans}}  & \parbox[t]{2mm}{\rotatebox[origin=c]{90}{ BLIVA}}  & \parbox[t]{2mm}{\rotatebox[origin=c]{90}{ VisualGLM}} &  Avg.$\Delta$\\\midrule
                          & Zero-Shot & 87.35  & 88.11  & 82.41  & 84.63  & 89.23  & 91.81  & \color{gray}{0.00} \\\midrule
        \multirow{3}{*}{MiniGPT-4} & VP~\cite{bahng2022exploring} & 92.40$^\ast$ & 82.07 & 78.58 & 85.82 & 78.85 & 81.29 & -4.09\\
                          & EVP~\cite{wu2022unleashing}  & 97.97$^\ast$  & 84.57 & 83.39 & 86.93 & 86.45 & 85.92 & +0.28 \\
                          &TVP (ours)   & \bf{98.33}$^\ast$ & 92.82  & 91.69 & 88.70 & 87.48 & 87.53 &  {\color{purple}\bf{+3.83}}\\\midrule
        \multirow{3}{*}{InstructBLIP} & VP~\cite{bahng2022exploring} & 82.80 & 91.22$^\ast$ & 87.46 & 83.71 & 85.92 & 85.21 & -1.20 \\
                          & EVP~\cite{wu2022unleashing}  & 87.52 & 97.81$^\ast$ & 86.18 & 87.16 & 94.39 & 90.28 & +3.30 \\
                          &TVP (ours)   & 91.69  & \bf{98.07}$^\ast$ & \bf{96.02} & 91.09 & \bf{97.78} & 89.01 & {\color{purple}\bf{+6.69}}\\\midrule
        \multirow{3}{*}{Ensemble} & VP~\cite{bahng2022exploring} & 90.84$^\ast$ & 90.95$^\ast$ & 91.91 & 88.84 & 92.00 & 81.43 & +2.07\\
                          & EVP~\cite{wu2022unleashing}  & 97.69$^\ast$ & 97.63$^\ast$ & 89.40 & 91.43 & 95.47 & 88.29 & +6.06\\
                          &TVP (ours)   & 97.18$^\ast$ & 96.21$^\ast$  & 95.46 & \bf{92.92} & 96.20 & \bf{92.06} & {\color{purple}\bf{+7.75}}\\\bottomrule
        \end{tabular}}
    \end{subtable}%
    \hfill
    \begin{subtable}{.495\linewidth}  
        \centering
        \scriptsize
        \setlength{\tabcolsep}{3pt}
        \resizebox{\linewidth}{!}{
        \begin{tabular}{c|c|cccccc>{\columncolor[gray]{0.9}}c}
        \toprule\toprule
        \multicolumn{2}{c|}{\bf Recognition: ImageNette} & \parbox[t]{2mm}{\rotatebox[origin=c]{90}{ MiniGPT-4}}  & \parbox[t]{2mm}{\rotatebox[origin=c]{90}{ InstructBLIP}}  & \parbox[t]{2mm}{\rotatebox[origin=c]{90}{ BLIP2}}  & \parbox[t]{2mm}{\rotatebox[origin=c]{90}{ VPGTrans}}  & \parbox[t]{2mm}{\rotatebox[origin=c]{90}{ BLIVA}}  & \parbox[t]{2mm}{\rotatebox[origin=c]{90}{ VisualGLM}} &  Avg.$\Delta$\\\midrule
                          & Zero-Shot &  79.82  & 74.78  & 95.03  & 82.57  & 77.02  &73.20  & \color{gray}{0.00} \\\midrule
        \multirow{3}{*}{MiniGPT-4} & VP~\cite{bahng2022exploring} & 83.39$^\ast$ & 71.12 & \bf{95.59} & 81.40 & 80.33 & 73.73 & +0.52 \\
                          & EVP~\cite{wu2022unleashing}  & 96.79$^\ast$ & 68.15 & 91.36 & 79.08 & 75.82 & 76.05 & +0.81\\
                          &TVP (ours)   & \bf{97.71}$^\ast$ & 78.34 & 94.98 & {86.34} & 84.51 & 75.34 & {\color{purple}\bf{+5.80}}\\\midrule
        \multirow{3}{*}{InstructBLIP} & VP~\cite{bahng2022exploring} & 85.50 & 93.22$^\ast$ & 90.88 & 79.67 & 92.94 & {76.71} & +6.08 \\
                          & EVP~\cite{wu2022unleashing}  & 84.05 & 96.92$^\ast$ & 94.65 & 80.87 & 90.11 & 76.46 & +6.77 \\
                          &TVP (ours)   & 85.58 & \bf{98.24}$^\ast$ & 92.71 & 83.95 & \bf{96.79} & \bf{80.46} & {\color{purple}\bf{+9.22}} \\\midrule
        \multirow{3}{*}{Ensemble} & VP~\cite{bahng2022exploring} & 85.48$^\ast$ & 86.55$^\ast$ & 93.99 & 79.08 & 86.27 & 71.77  & +3.45 \\
                          & EVP~\cite{wu2022unleashing}  & {97.63}$^\ast$ & {97.16}$^\ast$ & 92.37 & 83.19 & 90.04 & 76.49 & +9.08\\
                          &TVP (ours)   & 97.22$^\ast$ & 97.00$^\ast$ & 91.54 & \bf{92.60} & {94.37} & 76.69 & {\color{purple}\bf{+11.17}}\\\bottomrule
        \end{tabular}}
    \end{subtable}

    \begin{subtable}{.495\linewidth}  
        \centering
        \scriptsize
        \setlength{\tabcolsep}{3pt}
        \resizebox{\linewidth}{!}{
        \begin{tabular}{c|c|cccccc>{\columncolor[gray]{0.9}}c}
        \toprule
        \multicolumn{2}{c|}{\bf Recognition: SVHN} & \parbox[t]{2mm}{\rotatebox[origin=c]{90}{ MiniGPT-4}}  & \parbox[t]{2mm}{\rotatebox[origin=c]{90}{ InstructBLIP}}  & \parbox[t]{2mm}{\rotatebox[origin=c]{90}{ BLIP2}}  & \parbox[t]{2mm}{\rotatebox[origin=c]{90}{ VPGTrans}}  & \parbox[t]{2mm}{\rotatebox[origin=c]{90}{ BLIVA}}  & \parbox[t]{2mm}{\rotatebox[origin=c]{90}{ VisualGLM}} &  Avg.$\Delta$\\\midrule
                          & Zero-Shot & 38.82  & 28.96  & 33.14  & 31.58  & 33.32  & 20.87  & \color{gray}{0.00} \\\midrule
        \multirow{3}{*}{MiniGPT-4} & VP~\cite{bahng2022exploring} & 65.58$^\ast$ & 37.80 & 52.66 & 42.97 & 27.02 & 20.12 & +9.91 \\
                          & EVP~\cite{wu2022unleashing}  & 74.24$^\ast$ & 41.59 & 48.87 & \bf{57.61} & 36.11 & \bf{33.12} & +17.48\\
                          &TVP (ours)   & {75.17}$^\ast$ & 54.32 & 61.95 & {51.10} & 60.28 & {32.17} &  {\color{purple}\bf{+24.72}}\\\midrule
        \multirow{3}{*}{InstructBLIP} & VP~\cite{bahng2022exploring} & 29.53 & 73.20$^\ast$ & 51.63 & 36.82 & 50.98 & 21.86 & +12.89 \\
                          & EVP~\cite{wu2022unleashing}  & 39.80 & {87.55}$^\ast$ & 45.24 & 31.37 & 44.74 & 17.34 & +13.23\\
                          &TVP (ours)   & 54.37 & \bf{89.87}$^\ast$ & {70.61} & 46.82 & {70.92} & 25.38  &  {\color{purple}\bf{+28.55}}\\\midrule
        \multirow{3}{*}{Ensemble} & VP~\cite{bahng2022exploring} & 58.41$^\ast$ & 64.65$^\ast$ & 47.95 & 47.25 & 50.79 & 25.00 & +17.89\\
                          & EVP~\cite{wu2022unleashing}  & 70.14$^\ast$ & 70.23$^\ast$ & 53.27 & 29.08 & 65.66 & 22.35 &  +20.67\\
                          &TVP (ours)   & \bf{82.58}$^\ast$ & 81.82$^\ast$ & \bf{73.06} & 37.09 & \bf{73.46} & 28.61 &  {\color{purple}\bf{+31.54}}\\\bottomrule
        \end{tabular}}
    \end{subtable}%
    \hfill
    \begin{subtable}{.495\linewidth}  
        \centering
        \scriptsize
        \setlength{\tabcolsep}{3pt}
        \resizebox{\linewidth}{!}{
        \begin{tabular}{c|c|cccccc>{\columncolor[gray]{0.9}}c}
        \toprule
        \multicolumn{2}{c|}{\bf Counting: CLEVR} & \parbox[t]{2mm}{\rotatebox[origin=c]{90}{ MiniGPT-4}}  & \parbox[t]{2mm}{\rotatebox[origin=c]{90}{ InstructBLIP}}  & \parbox[t]{2mm}{\rotatebox[origin=c]{90}{ BLIP2}}  & \parbox[t]{2mm}{\rotatebox[origin=c]{90}{ VPGTrans}}  & \parbox[t]{2mm}{\rotatebox[origin=c]{90}{ BLIVA}}  & \parbox[t]{2mm}{\rotatebox[origin=c]{90}{ VisualGLM}} &  Avg.$\Delta$\\\midrule
                          & Zero-Shot &9.50  & 40.33  & 12.73  & 13.13  & 26.27  & 13.43  & \color{gray}{0.00} \\\midrule
        \multirow{3}{*}{MiniGPT-4} & VP~\cite{bahng2022exploring} & 35.73$^\ast$ & 33.57 & 12.77 & 8.80 & 29.07 & 12.77 & +2.89\\
                          & EVP~\cite{wu2022unleashing}  & {52.17}$^\ast$ & 39.03& 20.17 & 8.00 & 34.23 & 13.60 & +8.64 \\
                          &TVP (ours)   & 51.00$^\ast$ & 42.90 & 22.07 & 19.50 & 36.00 & 13.00 & {\color{purple}\bf{+11.51}} \\\midrule
        \multirow{3}{*}{InstructBLIP} & VP~\cite{bahng2022exploring} & 12.53 & 56.27$^\ast$ & 15.57 & 21.00 & {44.20} & {18.03} & +8.70 \\
                          & EVP~\cite{wu2022unleashing}  & 12.73 & {61.93}$^\ast$ & 13.40 & 10.40 & 39.90 & 13.87 & +6.14\\
                          &TVP (ours)   & 21.80  & \bf{63.60}$^\ast$ & \bf{33.73} & \bf{26.27} & \bf{46.93} & \bf{35.63} & {\color{purple}\bf{+18.76}}\\\midrule
        \multirow{3}{*}{Ensemble} & VP~\cite{bahng2022exploring} & 35.03$^\ast$ & 39.43$^\ast$ & 16.70 & 20.83 & 40.43 & 13.03 & +8.34\\
                          & EVP~\cite{wu2022unleashing}  & 46.33$^\ast$ & 54.63$^\ast$ & 24.87 & {23.77} & 38.27 & 13.43 & +14.22\\
                          &TVP (ours)   & \bf{53.60}$^\ast$ & 58.33$^\ast$ & {25.50} & 23.63 & 39.53 & 14.36 & {\color{purple}\bf{+16.49}}\\\bottomrule
        \end{tabular}}
    \end{subtable}

    \begin{subtable}{.495\linewidth}  
        \centering
        \scriptsize
        \setlength{\tabcolsep}{3pt}
        \resizebox{\linewidth}{!}{
        \begin{tabular}{c|c|cccccc>{\columncolor[gray]{0.9}}c}
        \toprule
        \multicolumn{2}{c|}{\bf Reasoning: HM} & \parbox[t]{2mm}{\rotatebox[origin=c]{90}{ MiniGPT-4}}  & \parbox[t]{2mm}{\rotatebox[origin=c]{90}{ InstructBLIP}}  & \parbox[t]{2mm}{\rotatebox[origin=c]{90}{ BLIP2}}  & \parbox[t]{2mm}{\rotatebox[origin=c]{90}{ VPGTrans}}  & \parbox[t]{2mm}{\rotatebox[origin=c]{90}{ BLIVA}}  & \parbox[t]{2mm}{\rotatebox[origin=c]{90}{ VisualGLM}} &  Avg.$\Delta$\\\midrule
                          & Zero-Shot & 53.37  & 61.62  &  53.44 & 59.54  & 62.28  & 57.55  & \color{gray}{0.00} \\\midrule
        \multirow{3}{*}{MiniGPT-4} & VP~\cite{bahng2022exploring} & 55.55$^\ast$ & 57.26 & 55.54 & 55.47 & 56.05 & 52.73 & -2.53\\
                          & EVP~\cite{wu2022unleashing}  & 57.58$^\ast$ & 60.66 & 55.34 & 56.87 & 60.64 & 57.27 & +0.09\\
                          &TVP (ours)   & 56.93$^\ast$ & 62.38 & 56.20 & 60.19 & {64.09} & 58.15 & {\color{purple}\bf{+1.69}}\\\midrule
        \multirow{3}{*}{InstructBLIP} & VP~\cite{bahng2022exploring} & 53.20 & 61.73$^\ast$ & {56.24} & 61.53 & 62.22 & 56.26 & +0.56 \\
                          & EVP~\cite{wu2022unleashing}  & 55.78 & \bf{63.05}$^\ast$ & 51.19 & 60.98 & 61.67 & 58.00 & +0.48 \\
                          &TVP (ours)   & 55.65 & 62.54$^\ast$ & 55.52 & {61.61} & 62.77 & \bf{59.01} & {\color{purple}\bf{+1.55}}\\\midrule
        \multirow{3}{*}{Ensemble} & VP~\cite{bahng2022exploring} & 54.81$^\ast$ & 61.88$^\ast$ & 55.51 & 61.54 & 62.39 & 57.66 & +1.00\\
                          & EVP~\cite{wu2022unleashing}  & {58.85}$^\ast$ & 62.26$^\ast$ & 54.09 & 59.21 & 62.10 & 58.22 & +1.16\\
                          &TVP (ours)   & \bf{62.33}$^\ast$ & {62.77}$^\ast$ & \bf{58.77} & \bf{62.23} & \bf{64.67} & {58.47} &  {\color{purple}\bf{+3.57}}\\\bottomrule\bottomrule
        \end{tabular}}
    \end{subtable}%
    \hfill
    \begin{subtable}{.495\linewidth}  
        \centering
        \scriptsize
        \setlength{\tabcolsep}{3pt}
        \resizebox{\linewidth}{!}{
        \begin{tabular}{c|c|cccccc>{\columncolor[gray]{0.9}}c}
        \toprule
        \multicolumn{2}{c|}{\bf Hallucination: POPE} & \parbox[t]{2mm}{\rotatebox[origin=c]{90}{ MiniGPT-4}}  & \parbox[t]{2mm}{\rotatebox[origin=c]{90}{ InstructBLIP}}  & \parbox[t]{2mm}{\rotatebox[origin=c]{90}{ BLIP2}}  & \parbox[t]{2mm}{\rotatebox[origin=c]{90}{ VPGTrans}}  & \parbox[t]{2mm}{\rotatebox[origin=c]{90}{ BLIVA}}  & \parbox[t]{2mm}{\rotatebox[origin=c]{90}{ VisualGLM}} &  Avg.$\Delta$\\\midrule
                          & Zero-Shot &  53.07  & 73.87  & 49.80  & 58.60  & 78.07  &70.87  & \color{gray}{0.00} \\\midrule
        \multirow{3}{*}{MiniGPT-4} & VP~\cite{bahng2022exploring} & 53.87$^\ast$ & 69.27 & 50.00 & 61.07 & 72.13 & 69.80 & -1.36 \\
                          & EVP~\cite{wu2022unleashing}  & {68.06}$^\ast$ & 69.80 & 50.00 & 61.07 & 71.33 & 69.40 & +0.90\\
                          &TVP (ours)   & \bf{68.73}$^\ast$ & 75.13 & {51.40} & 64.47 & 72.67 & 71.00 & {\color{purple}\bf{+3.19}}\\\midrule
        \multirow{3}{*}{InstructBLIP} & VP~\cite{bahng2022exploring} & 50.67 & 77.33$^\ast$ & 50.00 & 60.87 & 74.40 & 67.47 & -0.59\\
                          & EVP~\cite{wu2022unleashing}  & 51.20 & \bf{79.33}$^\ast$ & 50.00 & 63.47 & 72.87 & 69.87 & +0.41 \\
                          &TVP (ours)   & 54.53 & {77.80}$^\ast$ & 50.20 & \bf{64.93} & 75.20 & 70.47 & {\color{purple}\bf{+1.48}} \\\midrule
        \multirow{3}{*}{Ensemble} & VP~\cite{bahng2022exploring} & 58.33$^\ast$ & 76.73$^\ast$ & 50.60 & {64.87} & 72.13 & {71.07} & +1.57 \\
                          & EVP~\cite{wu2022unleashing}  & 64.67$^\ast$ & 75.87$^\ast$ & 50.00 & 64.07 & 71.80 & 62.13 & +0.71\\
                          &TVP (ours)   & 59.13$^\ast$ & 77.53$^\ast$ & \bf{52.40} & \bf{64.93} & \bf{75.27} & \bf{72.40} & {\color{purple}\bf{+2.90}}\\\bottomrule\bottomrule
        \end{tabular}}
    \end{subtable}
    \caption{Results on 6 datasets of different tasks from object recognition and counting to multimodal reasoning and hallucination. Visual prompts are trained on MiniGPT-4~\cite{zhu2023minigpt}, InstructBLIP~\cite{Dai2023InstructBLIP} and their ensemble with different methods, and further tested on 6 modern MLLMs. We display the average improvements over all models on the last column and $\ast$ denotes the results of the model for prompt training. AUC score is the metric for Hatefulmemes (HM) and top-1 accuracy (\%) is used for the rest. Besides highlighting the best overall average improvement, we mark the best result in {\bf bold} for each model.}
    \vspace{-2ex}
\label{tab:main_results}
\end{table*}

\subsection{Experimental Settings}
Here, we briefly list the basic settings for the following experiments. More details are provided in~\cref{sec:implementation}.

\textbf{Datasets and Metrics.} We consider 10 datasets involving diverse visual or multimodal tasks. For visual tasks, we take 8 datasets of object recognition (\eg, CIFAR-10~\cite{krizhevsky2009learning}, ImageNette~\cite{imagenette} and SVHN~\cite{svhn}) and object counting (\eg, CLEVR~\cite{johnson2017clevr}) for illustration. We further focus on two challenging multimodal problems including multimodal reasoning (Hatefulmemes~\cite{kiela2020hateful}) and hallucination (POPE~\cite{li2023evaluating}), to better demonstrate the effectiveness of our methods in the realm of MLLM. AUC score is taken as the metric for Hatefulmemes while top-1 accuracy is used for the rest.


\textbf{Models.} We select 6 modern MLLMs with diverse capabilities as evaluated by~\cite{fu2023mme} to demonstrate that the generated visual prompts by our method can universally elevate their performance. To be specific, we take MiniGPT-4~\cite{zhu2023minigpt} and InstructBLIP~\cite{Dai2023InstructBLIP} for training visual prompts respectively and transfer the prompts to BLIP2~\cite{li2023blip}, VPGTrans~\cite{zhang2023transfer}, BLIVA~\cite{hu2023bliva} and VisualGLM~\cite{visualglm}.

\textbf{Baselines.} We mainly focus on the transferability of visual prompting across diverse models and compare the proposed TVP method with VP~\cite{bahng2022exploring} and EVP~\cite{wu2022unleashing}, which are general methods of visual prompting, rather than those focusing on generalization across different data distributions.

\textbf{Implementations.} Hyperparameters for TVP include two balance weights for the proposed loss terms. Following~\cite{jia2022vpt,oh2023blackvip}, we set them optimal by grid-search within small ranges on validation sets. The search ranges and other details are introduced in~\cref{sec:hyper}.

\subsection{Main Results}
\label{sec:main_results}

We train visual prompts on either MiniGPT-4 or InstructBLIP for target downstream tasks using different methods and examine their performance on the 6 selected models. The results are displayed in~\cref{tab:main_results}. Several findings are summarized in the following context. 

First of all, this is the first time that visual prompting technique~\cite{bahng2022exploring} is proved to be effective for MLLMs on multimodal tasks involving reasoning besides recognition. Beyond that, VP~\cite{bahng2022exploring} yields the least favorable outcomes, often leading to an overall performance degradation, indicating poorer transferability. EVP brings more significant improvements for the models training visual prompts due to the stable optimization with normalized gradients, which is aligned with~\cite{wu2022unleashing}, but its benefits for other models are limited.
In contrast, while the proposed TVP achieves similar effectiveness on models for training, it also boosts the performance of other models with larger margins. 
The overall effectiveness is manifested as the highest average delta in growth across 6 models on all downstream tasks, which better achieves our goal of enhancing different models with a single set of shared external parameters. 

Moreover, we have some more in-depth findings regarding the results of TVP. The improvements of VisualGLM~\cite{visualglm} are relatively modest compared to other models. This suggests that the visual prompts are more challenging to transfer to VisualGLM. The language model of VisualGLM is GLM~\cite{du2021glm} while the language models for the rest models are based on the LLaMA architecture~\cite{touvron2023llama}. This difference in language models can raise the difficulty of effective prompt transferring. Also, we notice that the transferability of visual prompts generated by InstructBLIP is better than those from MiniGPT-4 in general. Though sharing similar architecture, InstructBLIP is tuned systematically on 13 vision-language datasets~\cite{Dai2023InstructBLIP}, which is far more plentiful than MiniGPT-4~\cite{zhu2023minigpt}. This means that InstructBLIP possesses more internal knowledge concerning visual-language tasks, which should help the learning of visual prompts and benefit other models with more improvements.

Similar conclusions are drawn on more practical datasets like CIFAR-100~\cite{krizhevsky2009learning}, Oxford-Pets~\cite{parkhi12pets}, FGVC-Aircraft~\cite{maji13fine-grained} and Food101~\cite{food101}, in~\cref{tab:addition_results}, \cref{sec:result}. 

\subsection{Model Ensemble}

Ensembling methods have been proven effective to enhance model generalization~\cite{ueda1996generalization,bian2021does} and transfer attack~\cite{dong2018boosting, chen2023rethinking,dong2023robust}. The idea of ensemble can also be applied to transferable visual prompting. The intuition is that if the visual prompts are effective for multiple models, it is more likely that they contain more shareable informative semantics and transfer better to other models.
We simply optimize the visual prompts by averaging over losses of different models, and here take MiniGPT-4 and InstructBLIP for ensemble.

The results of ensemble prompt learning are presented in~\cref{tab:main_results}. We can see that model ensembling is a general technique that can enhance the transferability of visual prompts trained with different methods. Meanwhile, the trend that TVP has the best performance remains unchanged and with the aid of ensemble, it achieves the best overall improvements on four tasks. 
However, more models bring more computational and storage overheads for ensemble. A balance between training cost and prompt transferability needs to be sought when adopting this technique.

\begin{figure}[t]
    \centering
    \includegraphics[width=.9\linewidth]{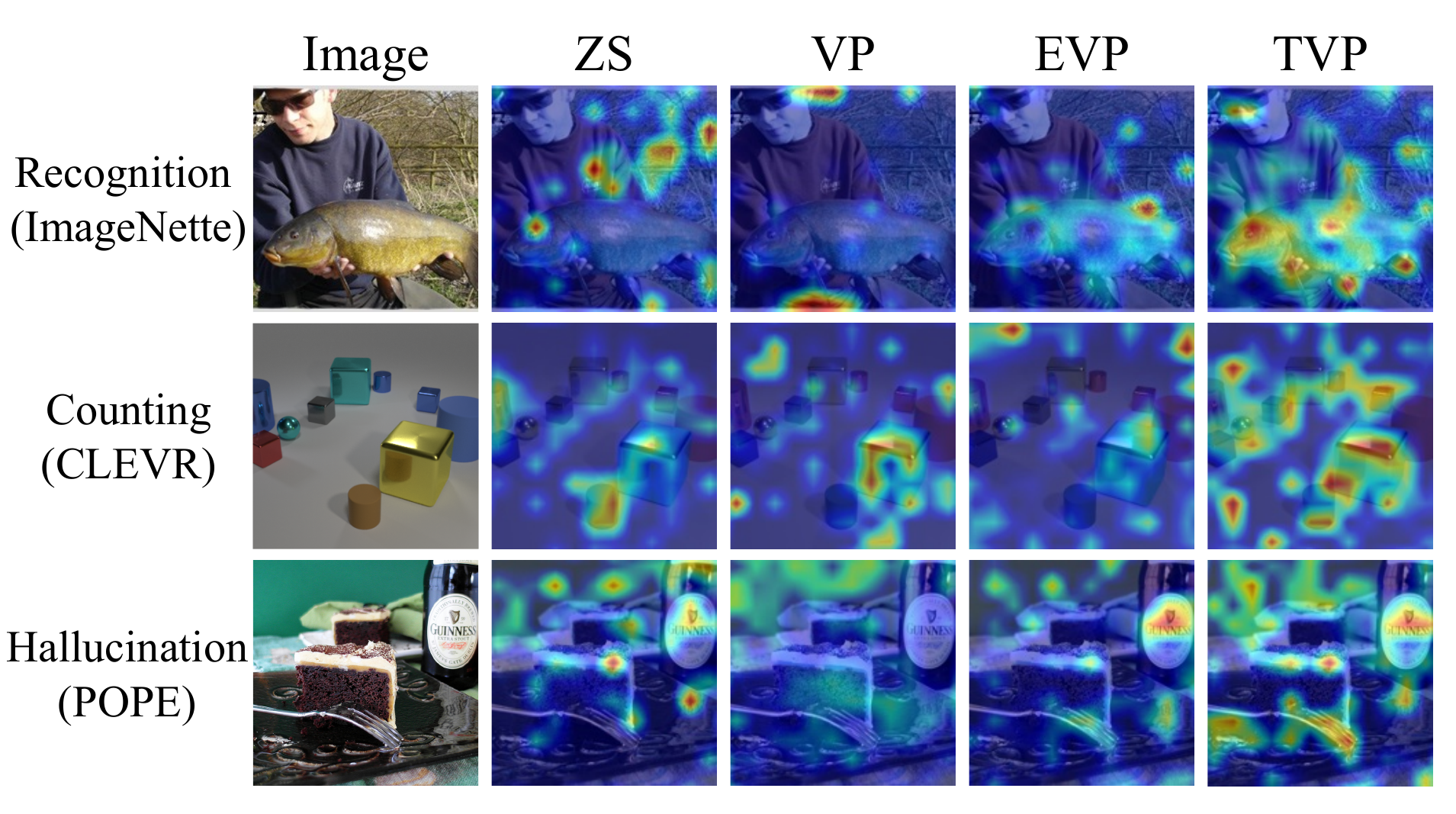}
    \vspace{-2ex}
    \caption{GradCAM~\cite{gradcam} of VPGTrans~\cite{zhang2023transfer} on 3 different tasks. TVP encourages the model to attend to task-related objects.}
    \vspace{-2ex}
    \label{fig:cam}
\end{figure}

\begin{table}[t]
    \centering
    \resizebox{\linewidth}{!}{
    \renewcommand{\arraystretch}{0.5}
    \begin{tabular}{cccccccc} 
    \toprule
    \toprule
         FCA&  TSE&  CIFAR-10 & IN &SVHN &  CLEVR &  HM & POPE\\ \hline
         \XSolidBrush &  \XSolidBrush& 0.28 & 0.81 & 17.45 & 8.64 & 0.09 & 0.90\\ 
         \Checkmark & \XSolidBrush & 3.81 & 2.73 & 20.84 & 10.83 & 0.99 & 2.33\\
        \XSolidBrush & \Checkmark & 1.91 & 3.36 & 24.68 & 10.14 & 0.31 & 2.15 \\  \hline
        \rowcolor[gray]{0.9}\Checkmark &  \Checkmark & \bf{3.83} & \bf{5.80} & \bf{24.72} & \bf{11.51} & \bf{1.69} & \bf{3.19} \\ \bottomrule\bottomrule
    \end{tabular}}
    \vspace{-1ex}
    \caption{The average performance improvements with different combinations of FCA and TSE.}
    \vspace{-2ex}
    \label{tab:ablation}
\end{table}

\begin{table}[t]
\centering
\resizebox{\linewidth}{!}{
\begin{tabular}{lccccccc}
\toprule\toprule
Prompt Width & 5 & 10 & 20 & 30 & 40 & 50 & 80\\
\midrule
MiniGPT-4 & +2.84  & +3.23  & \bf{+3.97}  & +3.83  & +2.71 & +1.93  & -3.03  \\
InstructBLIP & +3.71 & +5.42  & +5.06  & \bf{+6.69}  & +3.15 & +2.91  & -2.12  \\
Ensemble & +4.66 & +3.38 & +5.45 & \bf{+7.75} & +5.87 & +5.12 & +2.67 \\
\bottomrule\bottomrule
\end{tabular}
}
\vspace{-1ex}
\caption{The average improvements on CIFAR-10 with prompts of different widths by TVP using different training models.
}
\vspace{-2ex}
\label{tab:width}
\end{table}

\subsection{Ablation Studies}
\label{sec:ablation}
We conduct ablation studies on TVP to further verify our design. In the rest context, we only report average performance for analysis and detailed results are in~\cref{sec:result}.

\subsubsection{Strategies of FCA and TSE}
We first examine the impacts of the proposed strategies, FCA and TSE, on the performance of visual prompts trained with MiniGPT-4. 
As shown in~\cref{tab:ablation}, individually applying either strategy can improve the average performance to some extent. When they are combined together, it can maximize the assistance of visual prompts for diverse models. This indicates that both task-agnostic prior knowledge motivating FCA and task-related feature extraction enhanced by TSE can contribute to the transferability of visual prompts. Visualization in~\cref{fig:cam} also suggests that TVP can help models better locate the objects beneficial for task completion.

\subsubsection{Prompt Width}

Prompt width, defining the number of parameters, could be critical for TVP's performance. The results on CIFAR-10 in~\cref{tab:width} show that there is a trade-off between the number of learnable parameters and the scaled image size, and the performance of TVP reaches the peak at a moderate prompt width around 20-30, which validates our choice of 30.

\subsection{In-depth Analyses}
\label{sec:analyses}
We present several in-depth analyses below, to demonstrate the effectiveness of TVP under different conditions and validate its practicality in real scenarios.

\begin{figure}[t]
    \centering
    \begin{subfigure}{.33\linewidth}
        \centering
        \includegraphics[trim={0 2cm 0 2cm},width=\linewidth]{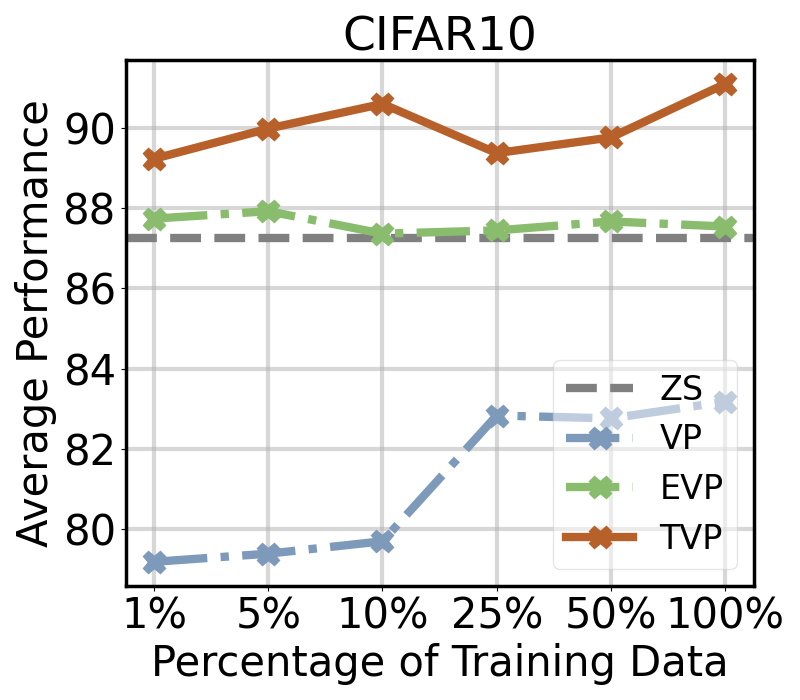}
        \label{fig:cifar}
    \end{subfigure}%
    \hfill
    \begin{subfigure}{.33\linewidth}
        \centering
        \includegraphics[trim={0 2cm 0 2cm},width=\linewidth]{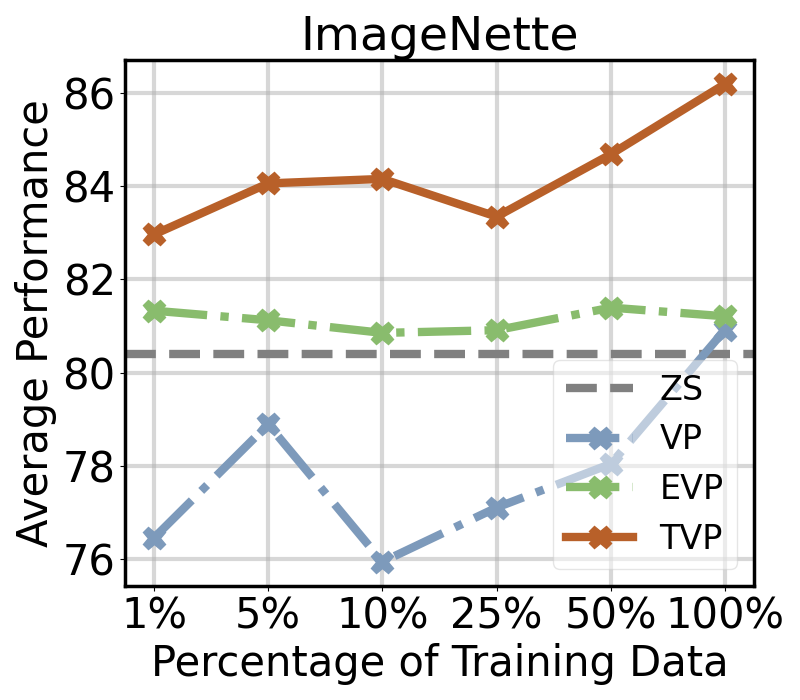}
        \label{fig:imagenette}
    \end{subfigure}
    \hfill
    \begin{subfigure}{.33\linewidth}
        \centering
        \includegraphics[trim={0 2cm 0 2cm},width=\linewidth]{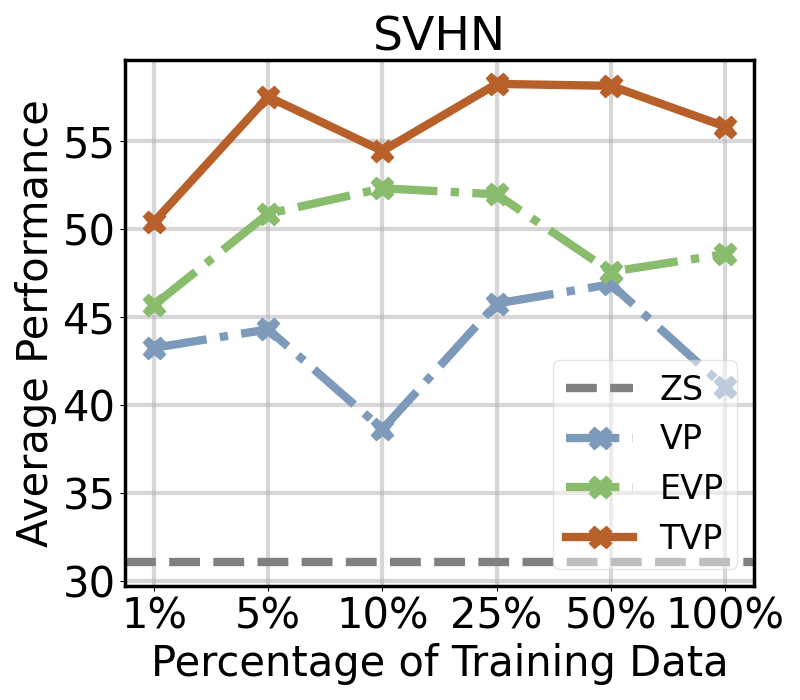}
        \label{fig:svhn}
    \end{subfigure}%
    \newline
    \begin{subfigure}{.33\linewidth}
        \centering
        \includegraphics[width=\linewidth]{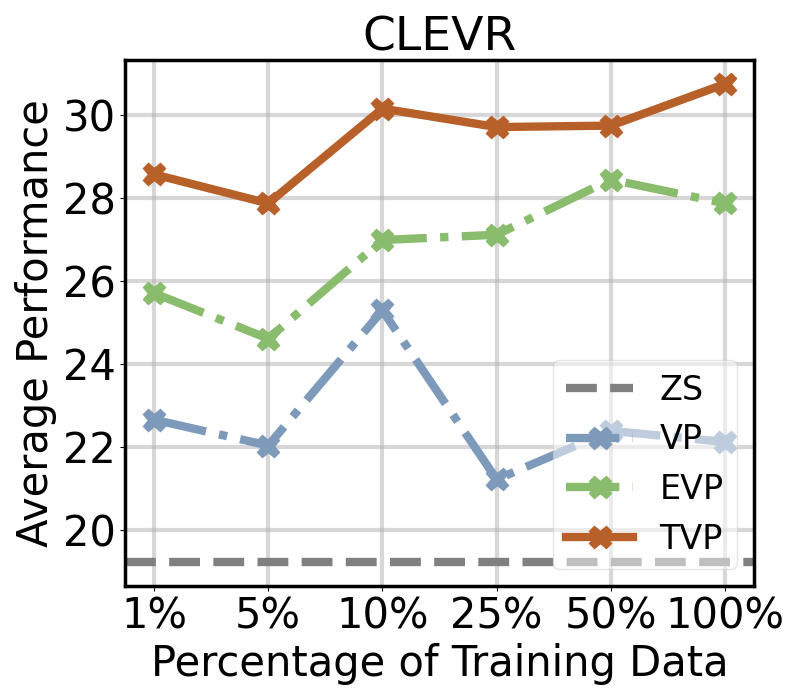}
        \label{fig:clevr}
    \end{subfigure}%
    \hfill
    \begin{subfigure}{.33\linewidth}
        \centering
        \includegraphics[width=\linewidth]{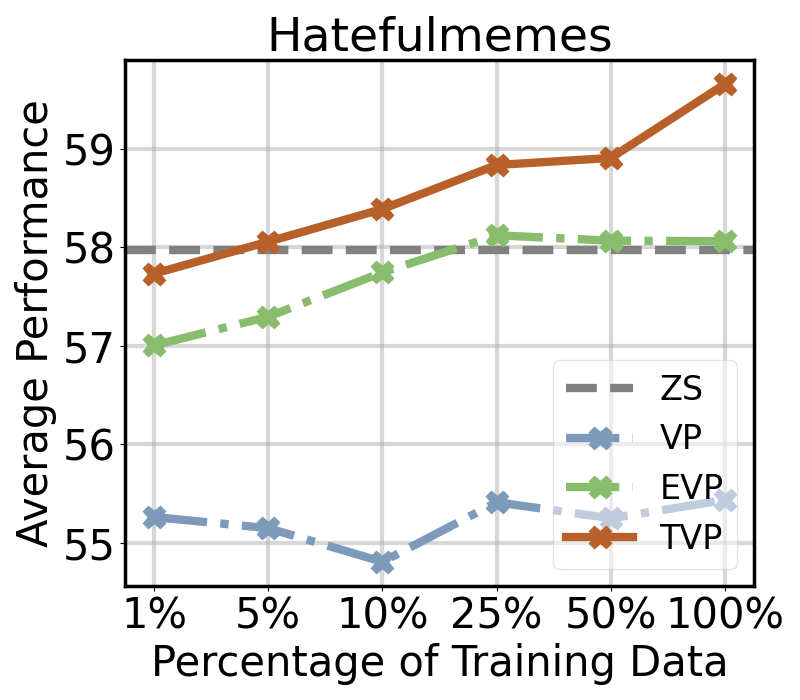}
        \label{fig:hateful}
    \end{subfigure}
    \hfill
    \begin{subfigure}{.33\linewidth}
        \centering
        \includegraphics[width=\linewidth]{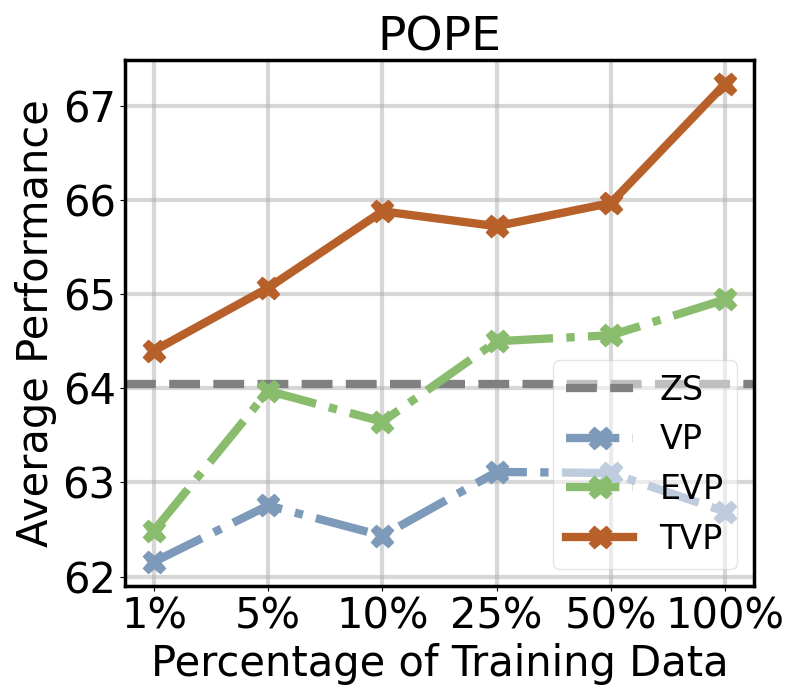}
        \label{fig:pope}
    \end{subfigure}%
    \vspace{-4ex}
    \caption{Curves of average performance as the training data scale changes. TVP can effectively enhance the performance of different models even with only 1\% of the data, and its overall performance improves as the data size increases.}
    \vspace{-2ex}
    \label{fig:data-scale}
\end{figure}
\vspace{-2ex}
\subsubsection{Data Scale}

Considering that the data available for adaptation is sometimes limited in practical scenarios, it's essential to study the impact of the training data size on the performance of the learned visual prompts. 
Following~\cite{jia2022vpt,wu2022unleashing}, we examine the varying trends of different methods within a larger range from 1\% to 100\%. We plot the curves of the average performance of different models armed with the visual prompts trained using MiniGPT-4 as the amount of data changes in~\cref{fig:data-scale}. As expected, the performance of TVP tends to improve as the data scale increases in general. The overall conclusion in~\cref{sec:main_results} that TVP performs better than VP and EVP remains consistent regardless of data sizes. It is worth noting that TVP can enhance the overall performance of multiple models even when the data amount is only 1\%, and the effect of visual prompts is sometimes close to that trained with the complete datasets.
This further illustrates the effectiveness of our method and makes its application in real-world scenarios more promising.

\vspace{-2ex}
\subsubsection{Generalization across Datasets}

It is worth investigating the generalization of trained visual prompts to further confirm the practicality of TVP in real scenarios. We take object recognition as an example.
Following the common practice in prompt learning~\cite{zhou2022conditional,khattak2023maple}, we apply the visual prompts generated with ensemble method on CIFAR-100, which has the most categories for common objects, to other datasets of the recognition task. As shown in~\cref{tab:generalization}, besides the best transferability on source dataset, TVP also gets the most or comparable improvements on other datasets, showing good generalization across diverse datasets within the same task.

\begin{table}[t]
    \centering
    \resizebox{\linewidth}{!}{
    \renewcommand{\arraystretch}{0.8}
    \begin{tabular}{l|c|cccccc}
    \toprule\toprule
    Avg. $\Delta$     & C-100 & C-10 & IN & SVHN & Pet & Air & Food \\\midrule
    VP   &   +2.87& +1.57& +1.24& -0.29& +2.87& \bf{+0.97}& -1.04\\
    EVP  &   +6.21& -0.11& +0.14& -2.84& +1.21& +0.75& -1.05\\
    TVP  &   \bf{+7.57}& \bf{+4.10}& \bf{+2.33}& \bf{+1.57}& \bf{+3.87}& +0.35& \bf{+1.89}\\
    \bottomrule\bottomrule
    \end{tabular}}
    \vspace{-1ex}
    \caption{Cross-dataset generalization with visual prompts using the ensemble of MiniGPT-4 and InstructBLIP on CIFAR-100.}
    \vspace{-2ex}
    \label{tab:generalization}
\end{table}

\subsubsection{Robustness to Corruptions}

We also test the robustness of visual prompts to common image corruptions~\cite{hendrycks2019robustness}. The visual prompt generated by TVP on MiniGPT-4 gets an average improvement of 2.30\% on CIFAR-10-C while both VP and EVP result in performance degradation of -6.62\% and -3.87\%. Results on corrupted datasets are in~\cref{sec:corruption} due to space limit.

\subsection{Discussion on Computational Efficiency}

We further compare the performance and efficiency of TVP with those of other fine-tuning methods and visual prompting methods, to support the motivation of the proposed problem and the corresponding solution of TVP. Due to the page limit, we present detailed discussion in~\cref{sec:efficiency}.

\section{Conclusion}

In this paper, we propose to optimize a set of \textit{shared parameters} for diverse MLLMs to adapt them to downstream tasks in a resource-friendly and flexible manner, which can avoid the computation and storage overheads with model-specific fine-tuning. Concretely, we introduce \textbf{Transferable Visual Prompting} to boost the performance of a group of models by adopting visual prompts as the shared parameters and improving their transferability with one-time training on only one model. Existing methods of visual prompting usually fail to enhance unseen models by satisfying margins due to feature corruption. We address this with two key strategies, Feature Consistency Alignment and Task Semantics Enrichment, which maintain the inner prior knowledge of large-scale pre-trained models and strengthen the task-related features extracted by models. Through extensive experiments on 10 datasets of diverse tasks from recognition and counting to multimodal reasoning and hallucination correction, we demonstrate the effectiveness of the proposed TVP to promote different models simultaneously. 
\section*{Acknowledgement}
This work was supported by the National Key Research and Development Program of China (No. 2020AAA0106302), NSFC Projects (Nos.~92370124, 62350080, 92248303, 62276149, U2341228, 62061136001, 62076147), BNRist (BNR2022RC01006), Tsinghua Institute for Guo Qiang, and the High Performance Computing Center, Tsinghua University. J. Zhu was also supported by the XPlorer Prize. Y. Dong was also supported by the China National Postdoctoral Program for Innovative Talents and Shuimu Tsinghua Scholar Program. 
\normalsize

{
    \small
    \bibliographystyle{ieeenat_fullname}
    \bibliography{main}

\begin{thebibliography}{69}
\providecommand{\natexlab}[1]{#1}
\providecommand{\url}[1]{\texttt{#1}}
\expandafter\ifx\csname urlstyle\endcsname\relax
  \providecommand{\doi}[1]{doi: #1}\else
  \providecommand{\doi}{doi: \begingroup \urlstyle{rm}\Url}\fi

\bibitem[vis(2023)]{visualglm}
Visualglm-6b.
\newblock \url{https://github.com/THUDM/VisualGLM-6B/}, 2023.

\bibitem[Alayrac et~al.(2022)Alayrac, Donahue, Luc, Miech, Barr, Hasson, Lenc, Mensch, Millican, Reynolds, et~al.]{alayrac2022flamingo}
Jean-Baptiste Alayrac, Jeff Donahue, Pauline Luc, Antoine Miech, Iain Barr, Yana Hasson, Karel Lenc, Arthur Mensch, Katherine Millican, Malcolm Reynolds, et~al.
\newblock Flamingo: a visual language model for few-shot learning.
\newblock In \emph{Advances in Neural Information Processing Systems}, pages 23716--23736, 2022.

\bibitem[Bahng et~al.(2022)Bahng, Jahanian, Sankaranarayanan, and Isola]{bahng2022exploring}
Hyojin Bahng, Ali Jahanian, Swami Sankaranarayanan, and Phillip Isola.
\newblock Exploring visual prompts for adapting large-scale models.
\newblock \emph{arXiv preprint arXiv:2203.17274}, 2022.

\bibitem[Bian and Chen(2021)]{bian2021does}
Yijun Bian and Huanhuan Chen.
\newblock When does diversity help generalization in classification ensembles?
\newblock \emph{IEEE Transactions on Cybernetics}, 52\penalty0 (9):\penalty0 9059--9075, 2021.

\bibitem[Bommasani et~al.(2021)Bommasani, Hudson, Adeli, Altman, Arora, von Arx, Bernstein, Bohg, Bosselut, Brunskill, et~al.]{bommasani2021opportunities}
Rishi Bommasani, Drew~A Hudson, Ehsan Adeli, Russ Altman, Simran Arora, Sydney von Arx, Michael~S Bernstein, Jeannette Bohg, Antoine Bosselut, Emma Brunskill, et~al.
\newblock On the opportunities and risks of foundation models.
\newblock \emph{arXiv preprint arXiv:2108.07258}, 2021.

\bibitem[Bossard et~al.(2014)Bossard, Guillaumin, and Van~Gool]{food101}
Lukas Bossard, Matthieu Guillaumin, and Luc Van~Gool.
\newblock Food-101 -- mining discriminative components with random forests.
\newblock In \emph{Computer Vision -- ECCV 2014}, pages 446--461, 2014.

\bibitem[Chen et~al.(2023)Chen, Zhang, Dong, and Zhu]{chen2023rethinking}
Huanran Chen, Yichi Zhang, Yinpeng Dong, and Jun Zhu.
\newblock Rethinking model ensemble in transfer-based adversarial attacks.
\newblock \emph{arXiv preprint arXiv:2303.09105}, 2023.

\bibitem[Chiang et~al.(2023)Chiang, Li, Lin, Sheng, Wu, Zhang, Zheng, Zhuang, Zhuang, Gonzalez, Stoica, and Xing]{vicuna2023}
Wei-Lin Chiang, Zhuohan Li, Zi Lin, Ying Sheng, Zhanghao Wu, Hao Zhang, Lianmin Zheng, Siyuan Zhuang, Yonghao Zhuang, Joseph~E. Gonzalez, Ion Stoica, and Eric~P. Xing.
\newblock Vicuna: An open-source chatbot impressing gpt-4 with 90\%* chatgpt quality, 2023.

\bibitem[Dai et~al.(2023)Dai, Li, Li, Tiong, Zhao, Wang, Li, Fung, and Hoi]{Dai2023InstructBLIP}
Wenliang Dai, Junnan Li, Dongxu Li, Anthony Meng~Huat Tiong, Junqi Zhao, Weisheng Wang, Boyang Li, Pascale Fung, and Steven C.~H. Hoi.
\newblock Instructblip: Towards general-purpose vision-language models with instruction tuning.
\newblock \emph{arXiv preprint arXiv:2305.06500}, 2023.

\bibitem[Dong et~al.(2018)Dong, Liao, Pang, Su, Zhu, Hu, and Li]{dong2018boosting}
Yinpeng Dong, Fangzhou Liao, Tianyu Pang, Hang Su, Jun Zhu, Xiaolin Hu, and Jianguo Li.
\newblock Boosting adversarial attacks with momentum.
\newblock In \emph{Proceedings of the IEEE/CVF Conference on Computer Vision and Pattern Recognition (CVPR)}, pages 9185--9193, 2018.

\bibitem[Dong et~al.(2019)Dong, Pang, Su, and Zhu]{dong2019evading}
Yinpeng Dong, Tianyu Pang, Hang Su, and Jun Zhu.
\newblock Evading defenses to transferable adversarial examples by translation-invariant attacks.
\newblock In \emph{Proceedings of the IEEE/CVF Conference on Computer Vision and Pattern Recognition (CVPR)}, pages 4312--4321, 2019.

\bibitem[Dong et~al.(2023)Dong, Chen, Chen, Fang, Yang, Zhang, Tian, Su, and Zhu]{dong2023robust}
Yinpeng Dong, Huanran Chen, Jiawei Chen, Zhengwei Fang, Xiao Yang, Yichi Zhang, Yu Tian, Hang Su, and Jun Zhu.
\newblock How robust is google's bard to adversarial image attacks?
\newblock \emph{arXiv preprint arXiv:2309.11751}, 2023.

\bibitem[Du et~al.(2022)Du, Qian, Liu, Ding, Qiu, Yang, and Tang]{du2021glm}
Zhengxiao Du, Yujie Qian, Xiao Liu, Ming Ding, Jiezhong Qiu, Zhilin Yang, and Jie Tang.
\newblock {GLM}: General language model pretraining with autoregressive blank infilling.
\newblock In \emph{Proceedings of the 60th Annual Meeting of the Association for Computational Linguistics (Volume 1: Long Papers)}, pages 320--335, 2022.

\bibitem[Elsayed et~al.(2019)Elsayed, Goodfellow, and Sohl-Dickstein]{elsayed2018adversarial}
Gamaleldin~F. Elsayed, Ian Goodfellow, and Jascha Sohl-Dickstein.
\newblock Adversarial reprogramming of neural networks.
\newblock In \emph{International Conference on Learning Representations}, 2019.

\bibitem[Fu et~al.(2023)Fu, Chen, Shen, Qin, Zhang, Lin, Qiu, Lin, Yang, Zheng, et~al.]{fu2023mme}
Chaoyou Fu, Peixian Chen, Yunhang Shen, Yulei Qin, Mengdan Zhang, Xu Lin, Zhenyu Qiu, Wei Lin, Jinrui Yang, Xiawu Zheng, et~al.
\newblock Mme: A comprehensive evaluation benchmark for multimodal large language models.
\newblock \emph{arXiv preprint arXiv:2306.13394}, 2023.

\bibitem[Gunjal et~al.(2023)Gunjal, Yin, and Bas]{gunjal2023detecting}
Anisha Gunjal, Jihan Yin, and Erhan Bas.
\newblock Detecting and preventing hallucinations in large vision language models.
\newblock \emph{arXiv preprint arXiv:2308.06394}, 2023.

\bibitem[Hendrycks and Dietterich(2019)]{hendrycks2019robustness}
Dan Hendrycks and Thomas Dietterich.
\newblock Benchmarking neural network robustness to common corruptions and perturbations.
\newblock In \emph{International Conference on Learning Representations}, 2019.

\bibitem[Houlsby et~al.(2019)Houlsby, Giurgiu, Jastrzebski, Morrone, De~Laroussilhe, Gesmundo, Attariyan, and Gelly]{houlsby2019parameter}
Neil Houlsby, Andrei Giurgiu, Stanislaw Jastrzebski, Bruna Morrone, Quentin De~Laroussilhe, Andrea Gesmundo, Mona Attariyan, and Sylvain Gelly.
\newblock Parameter-efficient transfer learning for nlp.
\newblock In \emph{Proceedings of the 36th International Conference on Machine Learning}, pages 2790--2799, 2019.

\bibitem[Howard()]{imagenette}
Jeremy Howard.
\newblock imagenette.
\newblock \url{"https://github.com/fastai/imagenette/"}.

\bibitem[Hu et~al.(2021)Hu, Shen, Wallis, Allen-Zhu, Li, Wang, Wang, and Chen]{hu2021lora}
Edward~J Hu, Yelong Shen, Phillip Wallis, Zeyuan Allen-Zhu, Yuanzhi Li, Shean Wang, Lu Wang, and Weizhu Chen.
\newblock Lora: Low-rank adaptation of large language models.
\newblock \emph{arXiv preprint arXiv:2106.09685}, 2021.

\bibitem[Hu et~al.(2023)Hu, Xu, Li, Li, Chen, and Tu]{hu2023bliva}
Wenbo Hu, Yifan Xu, Y Li, W Li, Z Chen, and Z Tu.
\newblock Bliva: A simple multimodal llm for better handling of text-rich visual questions.
\newblock \emph{arXiv preprint arXiv:2308.09936}, 2023.

\bibitem[Huang et~al.(2023)Huang, Dong, Chen, Zhang, Wang, Hua, and Yu]{huang2023diversity}
Qidong Huang, Xiaoyi Dong, Dongdong Chen, Weiming Zhang, Feifei Wang, Gang Hua, and Nenghai Yu.
\newblock Diversity-aware meta visual prompting.
\newblock In \emph{Proceedings of the IEEE/CVF Conference on Computer Vision and Pattern Recognition (CVPR)}, pages 10878--10887, 2023.

\bibitem[Jawahar et~al.(2012)Jawahar, Zisserman, Vedaldi, and Parkhi]{parkhi12pets}
C.~V. Jawahar, A. Zisserman, A. Vedaldi, and O.~M. Parkhi.
\newblock Cats and dogs.
\newblock In \emph{2012 IEEE Conference on Computer Vision and Pattern Recognition (CVPR)}, pages 3498--3505, 2012.

\bibitem[Jia et~al.(2022)Jia, Tang, Chen, Cardie, Belongie, Hariharan, and Lim]{jia2022vpt}
Menglin Jia, Luming Tang, Bor-Chun Chen, Claire Cardie, Serge Belongie, Bharath Hariharan, and Ser-Nam Lim.
\newblock Visual prompt tuning.
\newblock In \emph{Computer Vision -- ECCV 2022}, pages 709--727, 2022.

\bibitem[Johnson et~al.(2017)Johnson, Hariharan, Van Der~Maaten, Fei-Fei, Lawrence~Zitnick, and Girshick]{johnson2017clevr}
Justin Johnson, Bharath Hariharan, Laurens Van Der~Maaten, Li Fei-Fei, C Lawrence~Zitnick, and Ross Girshick.
\newblock Clevr: A diagnostic dataset for compositional language and elementary visual reasoning.
\newblock In \emph{Proceedings of the IEEE/CVF Conference on Computer Vision and Pattern Recognition (CVPR)}, pages 2901--2910, 2017.

\bibitem[Khandelwal et~al.(2022)Khandelwal, Weihs, Mottaghi, and Kembhavi]{khandelwal2022simple}
Apoorv Khandelwal, Luca Weihs, Roozbeh Mottaghi, and Aniruddha Kembhavi.
\newblock Simple but effective: Clip embeddings for embodied ai.
\newblock In \emph{Proceedings of the IEEE/CVF Conference on Computer Vision and Pattern Recognition (CVPR)}, pages 14829--14838, 2022.

\bibitem[Khattak et~al.(2023{\natexlab{a}})Khattak, Rasheed, Maaz, Khan, and Khan]{khattak2023maple}
Muhammad~Uzair Khattak, Hanoona Rasheed, Muhammad Maaz, Salman Khan, and Fahad~Shahbaz Khan.
\newblock Maple: Multi-modal prompt learning.
\newblock In \emph{Proceedings of the IEEE/CVF Conference on Computer Vision and Pattern Recognition (CVPR)}, pages 19113--19122, 2023{\natexlab{a}}.

\bibitem[Khattak et~al.(2023{\natexlab{b}})Khattak, Wasim, Naseer, Khan, Yang, and Khan]{khattak2023self}
Muhammad~Uzair Khattak, Syed~Talal Wasim, Muzammal Naseer, Salman Khan, Ming-Hsuan Yang, and Fahad~Shahbaz Khan.
\newblock Self-regulating prompts: Foundational model adaptation without forgetting.
\newblock In \emph{Proceedings of the IEEE/CVF International Conference on Computer Vision (ICCV)}, pages 15190--15200, 2023{\natexlab{b}}.

\bibitem[Kiela et~al.(2020)Kiela, Firooz, Mohan, Goswami, Singh, Ringshia, and Testuggine]{kiela2020hateful}
Douwe Kiela, Hamed Firooz, Aravind Mohan, Vedanuj Goswami, Amanpreet Singh, Pratik Ringshia, and Davide Testuggine.
\newblock The hateful memes challenge: Detecting hate speech in multimodal memes.
\newblock In \emph{Advances in Neural Information Processing Systems}, pages 2611--2624, 2020.

\bibitem[Kim et~al.(2021)Kim, Yoo, Park, Kim, and Lee]{kim2021selfreg}
Daehee Kim, Youngjun Yoo, Seunghyun Park, Jinkyu Kim, and Jaekoo Lee.
\newblock Selfreg: Self-supervised contrastive regularization for domain generalization.
\newblock In \emph{Proceedings of the IEEE/CVF International Conference on Computer Vision (ICCV)}, pages 9619--9628, 2021.

\bibitem[Krizhevsky et~al.(2009)Krizhevsky, Hinton, et~al.]{krizhevsky2009learning}
Alex Krizhevsky, Geoffrey Hinton, et~al.
\newblock Learning multiple layers of features from tiny images.
\newblock 2009.

\bibitem[Li et~al.(2023{\natexlab{a}})Li, Wang, Wang, Ge, Ge, and Shan]{li2023seed}
Bohao Li, Rui Wang, Guangzhi Wang, Yuying Ge, Yixiao Ge, and Ying Shan.
\newblock Seed-bench: Benchmarking multimodal llms with generative comprehension.
\newblock \emph{arXiv preprint arXiv:2307.16125}, 2023{\natexlab{a}}.

\bibitem[Li et~al.(2023{\natexlab{b}})Li, Wong, Zhang, Usuyama, Liu, Yang, Naumann, Poon, and Gao]{li2023llava}
Chunyuan Li, Cliff Wong, Sheng Zhang, Naoto Usuyama, Haotian Liu, Jianwei Yang, Tristan Naumann, Hoifung Poon, and Jianfeng Gao.
\newblock Llava-med: Training a large language-and-vision assistant for biomedicine in one day.
\newblock \emph{arXiv preprint arXiv:2306.00890}, 2023{\natexlab{b}}.

\bibitem[Li et~al.(2023{\natexlab{c}})Li, Li, Savarese, and Hoi]{li2023blip}
Junnan Li, Dongxu Li, Silvio Savarese, and Steven Hoi.
\newblock Blip-2: Bootstrapping language-image pre-training with frozen image encoders and large language models.
\newblock \emph{arXiv preprint arXiv:2301.12597}, 2023{\natexlab{c}}.

\bibitem[Li et~al.(2023{\natexlab{d}})Li, Du, Zhou, Wang, Zhao, and Wen]{li2023evaluating}
Yifan Li, Yifan Du, Kun Zhou, Jinpeng Wang, Wayne~Xin Zhao, and Ji-Rong Wen.
\newblock Evaluating object hallucination in large vision-language models.
\newblock \emph{arXiv preprint arXiv:2305.10355}, 2023{\natexlab{d}}.

\bibitem[Lin et~al.(2014)Lin, Maire, Belongie, Hays, Perona, Ramanan, Doll{\'a}r, and Zitnick]{mscoco}
Tsung-Yi Lin, Michael Maire, Serge Belongie, James Hays, Pietro Perona, Deva Ramanan, Piotr Doll{\'a}r, and C.~Lawrence Zitnick.
\newblock Microsoft coco: Common objects in context.
\newblock In \emph{Computer Vision -- ECCV 2014}, pages 740--755, 2014.

\bibitem[Liu et~al.(2023{\natexlab{a}})Liu, Li, Wu, and Lee]{liu2023visual}
Haotian Liu, Chunyuan Li, Qingyang Wu, and Yong~Jae Lee.
\newblock Visual instruction tuning.
\newblock In \emph{Advances in Neural Information Processing Systems}, 2023{\natexlab{a}}.

\bibitem[Liu et~al.(2022)Liu, Ji, Fu, Tam, Du, Yang, and Tang]{liu2022p}
Xiao Liu, Kaixuan Ji, Yicheng Fu, Weng Tam, Zhengxiao Du, Zhilin Yang, and Jie Tang.
\newblock P-tuning: Prompt tuning can be comparable to fine-tuning across scales and tasks.
\newblock In \emph{Proceedings of the 60th Annual Meeting of the Association for Computational Linguistics (Volume 2: Short Papers)}, pages 61--68, 2022.

\bibitem[Liu et~al.(2023{\natexlab{b}})Liu, Yao, Ton, Zhang, Cheng, Klochkov, Taufiq, and Li]{liu2023trustworthy}
Yang Liu, Yuanshun Yao, Jean-Francois Ton, Xiaoying Zhang, Ruocheng Guo~Hao Cheng, Yegor Klochkov, Muhammad~Faaiz Taufiq, and Hang Li.
\newblock Trustworthy llms: a survey and guideline for evaluating large language models' alignment.
\newblock \emph{arXiv preprint arXiv:2308.05374}, 2023{\natexlab{b}}.

\bibitem[Luo et~al.(2023)Luo, Zhou, Ren, Chen, Sun, and Ji]{luo2023cheap}
Gen Luo, Yiyi Zhou, Tianhe Ren, Shengxin Chen, Xiaoshuai Sun, and Rongrong Ji.
\newblock Cheap and quick: Efficient vision-language instruction tuning for large language models.
\newblock In \emph{Advances in Neural Information Processing Systems}, pages 29615--29627, 2023.

\bibitem[Maji et~al.(2013)Maji, Kannala, Rahtu, Blaschko, and Vedaldi]{maji13fine-grained}
S. Maji, J. Kannala, E. Rahtu, M. Blaschko, and A. Vedaldi.
\newblock Fine-grained visual classification of aircraft.
\newblock Technical report, 2013.

\bibitem[Netzer et~al.(2011)Netzer, Wang, Coates, Bissacco, Wu, and Ng]{svhn}
Yuval Netzer, Tao Wang, Adam Coates, Alessandro Bissacco, Bo Wu, and Andrew~Y. Ng.
\newblock Reading digits in natural images with unsupervised feature learning.
\newblock In \emph{NIPS Workshop on Deep Learning and Unsupervised Feature Learning 2011}, 2011.

\bibitem[Oh et~al.(2023)Oh, Hwang, Lee, Lim, Jung, Jung, Choi, and Song]{oh2023blackvip}
Changdae Oh, Hyeji Hwang, Hee-young Lee, YongTaek Lim, Geunyoung Jung, Jiyoung Jung, Hosik Choi, and Kyungwoo Song.
\newblock Blackvip: Black-box visual prompting for robust transfer learning.
\newblock In \emph{Proceedings of the IEEE/CVF Conference on Computer Vision and Pattern Recognition (CVPR)}, pages 24224--24235, 2023.

\bibitem[Radford et~al.(2021)Radford, Kim, Hallacy, Ramesh, Goh, Agarwal, Sastry, Askell, Mishkin, Clark, et~al.]{radford2021clip}
Alec Radford, Jong~Wook Kim, Chris Hallacy, Aditya Ramesh, Gabriel Goh, Sandhini Agarwal, Girish Sastry, Amanda Askell, Pamela Mishkin, Jack Clark, et~al.
\newblock Learning transferable visual models from natural language supervision.
\newblock In \emph{Proceedings of the 38th International Conference on Machine Learning}, pages 8748--8763, 2021.

\bibitem[Roumeliotis and Tselikas(2023)]{roumeliotis2023chatgpt}
Konstantinos~I Roumeliotis and Nikolaos~D Tselikas.
\newblock Chatgpt and open-ai models: A preliminary review.
\newblock \emph{Future Internet}, 15\penalty0 (6):\penalty0 192, 2023.

\bibitem[Selvaraju et~al.(2017)Selvaraju, Cogswell, Das, Vedantam, Parikh, and Batra]{gradcam}
Ramprasaath~R. Selvaraju, Michael Cogswell, Abhishek Das, Ramakrishna Vedantam, Devi Parikh, and Dhruv Batra.
\newblock Grad-cam: Visual explanations from deep networks via gradient-based localization.
\newblock In \emph{Proceedings of the IEEE International Conference on Computer Vision (ICCV)}, 2017.

\bibitem[Shen et~al.(2022)Shen, Li, Tan, Bansal, Rohrbach, Chang, Yao, and Keutzer]{shen2021much}
Sheng Shen, Liunian~Harold Li, Hao Tan, Mohit Bansal, Anna Rohrbach, Kai-Wei Chang, Zhewei Yao, and Kurt Keutzer.
\newblock How much can clip benefit vision-and-language tasks?
\newblock In \emph{International Conference on Learning Representations}, 2022.

\bibitem[Su et~al.(2023)Su, Lan, Li, Xu, Wang, and Cai]{su2023pandagpt}
Yixuan Su, Tian Lan, Huayang Li, Jialu Xu, Yan Wang, and Deng Cai.
\newblock Pandagpt: One model to instruction-follow them all.
\newblock \emph{arXiv preprint arXiv:2305.16355}, 2023.

\bibitem[Touvron et~al.(2023{\natexlab{a}})Touvron, Lavril, Izacard, Martinet, Lachaux, Lacroix, Rozi{\`e}re, Goyal, Hambro, Azhar, et~al.]{touvron2023llama}
Hugo Touvron, Thibaut Lavril, Gautier Izacard, Xavier Martinet, Marie-Anne Lachaux, Timoth{\'e}e Lacroix, Baptiste Rozi{\`e}re, Naman Goyal, Eric Hambro, Faisal Azhar, et~al.
\newblock Llama: Open and efficient foundation language models.
\newblock \emph{arXiv preprint arXiv:2302.13971}, 2023{\natexlab{a}}.

\bibitem[Touvron et~al.(2023{\natexlab{b}})Touvron, Martin, Stone, Albert, Almahairi, Babaei, Bashlykov, Batra, Bhargava, Bhosale, et~al.]{touvron2023llama2}
Hugo Touvron, Louis Martin, Kevin Stone, Peter Albert, Amjad Almahairi, Yasmine Babaei, Nikolay Bashlykov, Soumya Batra, Prajjwal Bhargava, Shruti Bhosale, et~al.
\newblock Llama 2: Open foundation and fine-tuned chat models.
\newblock \emph{arXiv preprint arXiv:2307.09288}, 2023{\natexlab{b}}.

\bibitem[Tsai et~al.(2020)Tsai, Chen, and Ho]{tsai2020transfer}
Yun-Yun Tsai, Pin-Yu Chen, and Tsung-Yi Ho.
\newblock Transfer learning without knowing: Reprogramming black-box machine learning models with scarce data and limited resources.
\newblock In \emph{Proceedings of the 37th International Conference on Machine Learning}, pages 9614--9624, 2020.

\bibitem[Ueda and Nakano(1996)]{ueda1996generalization}
Naonori Ueda and Ryohei Nakano.
\newblock Generalization error of ensemble estimators.
\newblock In \emph{Proceedings of International Conference on Neural Networks (ICNN'96)}, pages 90--95. IEEE, 1996.

\bibitem[Van~der Maaten and Hinton(2008)]{van2008visualizing}
Laurens Van~der Maaten and Geoffrey Hinton.
\newblock Visualizing data using t-sne.
\newblock \emph{Journal of machine learning research}, 9\penalty0 (11), 2008.

\bibitem[Van~Laarhoven(2017)]{van2017l2}
Twan Van~Laarhoven.
\newblock L2 regularization versus batch and weight normalization.
\newblock \emph{arXiv preprint arXiv:1706.05350}, 2017.

\bibitem[Wei et~al.(2022)Wei, Bosma, Zhao, Guu, Yu, Lester, Du, Dai, and Le]{weifinetuned}
Jason Wei, Maarten Bosma, Vincent Zhao, Kelvin Guu, Adams~Wei Yu, Brian Lester, Nan Du, Andrew~M Dai, and Quoc~V Le.
\newblock Finetuned language models are zero-shot learners.
\newblock In \emph{International Conference on Learning Representations}, 2022.

\bibitem[Wu et~al.(2022)Wu, Li, Wei, Wang, Yuille, Zhou, and Xie]{wu2022unleashing}
Junyang Wu, Xianhang Li, Chen Wei, Huiyu Wang, Alan Yuille, Yuyin Zhou, and Cihang Xie.
\newblock Unleashing the power of visual prompting at the pixel level.
\newblock \emph{arXiv preprint arXiv:2212.10556}, 2022.

\bibitem[Wu et~al.(2023{\natexlab{a}})Wu, Yu, Zhou, Huang, Sun, and Ji]{wu2023parameter}
Qiong Wu, Wei Yu, Yiyi Zhou, Shubin Huang, Xiaoshuai Sun, and Rongrong Ji.
\newblock Parameter and computation efficient transfer learning for vision-language pre-trained models.
\newblock In \emph{Advances in Neural Information Processing Systems}, pages 41034--41050, 2023{\natexlab{a}}.

\bibitem[Wu et~al.(2023{\natexlab{b}})Wu, Wen, Backes, Berrang, Humbert, Shen, and Zhang]{wu2023quantifying}
Yixin Wu, Rui Wen, Michael Backes, Pascal Berrang, Mathias Humbert, Yun Shen, and Yang Zhang.
\newblock Quantifying privacy risks of prompts in visual prompt learning.
\newblock \emph{arXiv preprint arXiv:2310.11970}, 2023{\natexlab{b}}.

\bibitem[Xu et~al.(2023)Xu, Shao, Zhang, Gao, Liu, Lei, Meng, Huang, Qiao, and Luo]{xu2023lvlm}
Peng Xu, Wenqi Shao, Kaipeng Zhang, Peng Gao, Shuo Liu, Meng Lei, Fanqing Meng, Siyuan Huang, Yu Qiao, and Ping Luo.
\newblock Lvlm-ehub: A comprehensive evaluation benchmark for large vision-language models.
\newblock \emph{arXiv preprint arXiv:2306.09265}, 2023.

\bibitem[Yin et~al.(2023)Yin, Fu, Zhao, Li, Sun, Xu, and Chen]{yin2023survey}
Shukang Yin, Chaoyou Fu, Sirui Zhao, Ke Li, Xing Sun, Tong Xu, and Enhong Chen.
\newblock A survey on multimodal large language models.
\newblock \emph{arXiv preprint arXiv:2306.13549}, 2023.

\bibitem[Zeng et~al.(2022)Zeng, Liu, Du, Wang, Lai, Ding, Yang, Xu, Zheng, Xia, et~al.]{zeng2022glm}
Aohan Zeng, Xiao Liu, Zhengxiao Du, Zihan Wang, Hanyu Lai, Ming Ding, Zhuoyi Yang, Yifan Xu, Wendi Zheng, Xiao Xia, et~al.
\newblock Glm-130b: An open bilingual pre-trained model.
\newblock \emph{arXiv preprint arXiv:2210.02414}, 2022.

\bibitem[Zhai et~al.(2023)Zhai, Tong, Li, Cai, Qu, Lee, and Ma]{zhai2023investigating}
Yuexiang Zhai, Shengbang Tong, Xiao Li, Mu Cai, Qing Qu, Yong~Jae Lee, and Yi Ma.
\newblock Investigating the catastrophic forgetting in multimodal large language models.
\newblock \emph{arXiv preprint arXiv:2309.10313}, 2023.

\bibitem[Zhang et~al.(2023)Zhang, Fei, Yao, Ji, Li, Liu, and Chua]{zhang2023transfer}
Ao Zhang, Hao Fei, Yuan Yao, Wei Ji, Li Li, Zhiyuan Liu, and Tat-Seng Chua.
\newblock Transfer visual prompt generator across llms.
\newblock In \emph{Advances in Neural Information Processing Systems}, 2023.

\bibitem[Zhao et~al.(2023)Zhao, Cai, Si, Ma, An, Chen, Liu, Wang, Han, and Chang]{zhao2023mmicl}
Haozhe Zhao, Zefan Cai, Shuzheng Si, Xiaojian Ma, Kaikai An, Liang Chen, Zixuan Liu, Sheng Wang, Wenjuan Han, and Baobao Chang.
\newblock Mmicl: Empowering vision-language model with multi-modal in-context learning.
\newblock \emph{arXiv preprint arXiv:2309.07915}, 2023.

\bibitem[Zhou et~al.(2022{\natexlab{a}})Zhou, Yang, Loy, and Liu]{zhou2022conditional}
Kaiyang Zhou, Jingkang Yang, Chen~Change Loy, and Ziwei Liu.
\newblock Conditional prompt learning for vision-language models.
\newblock In \emph{Proceedings of the IEEE/CVF Conference on Computer Vision and Pattern Recognition (CVPR)}, pages 16816--16825, 2022{\natexlab{a}}.

\bibitem[Zhou et~al.(2022{\natexlab{b}})Zhou, Yang, Loy, and Liu]{zhou2022learning}
Kaiyang Zhou, Jingkang Yang, Chen~Change Loy, and Ziwei Liu.
\newblock Learning to prompt for vision-language models.
\newblock \emph{International Journal of Computer Vision}, 130\penalty0 (9):\penalty0 2337--2348, 2022{\natexlab{b}}.

\bibitem[Zhou et~al.(2023)Zhou, Liu, Qiao, Xiang, and Loy]{zhou2022domain}
Kaiyang Zhou, Ziwei Liu, Yu Qiao, Tao Xiang, and Chen~Change Loy.
\newblock Domain generalization: A survey.
\newblock \emph{IEEE Transactions on Pattern Analysis and Machine Intelligence}, 45\penalty0 (4):\penalty0 4396--4415, 2023.

\bibitem[Zhou et~al.(2018)Zhou, Hou, Chen, Tang, Huang, Gan, and Yang]{zhou2018transferable}
Wen Zhou, Xin Hou, Yongjun Chen, Mengyun Tang, Xiangqi Huang, Xiang Gan, and Yong Yang.
\newblock Transferable adversarial perturbations.
\newblock In \emph{Computer Vision -- ECCV 2018}, pages 452--467, 2018.

\bibitem[Zhu et~al.(2023)Zhu, Chen, Shen, Li, and Elhoseiny]{zhu2023minigpt}
Deyao Zhu, Jun Chen, Xiaoqian Shen, Xiang Li, and Mohamed Elhoseiny.
\newblock Minigpt-4: Enhancing vision-language understanding with advanced large language models.
\newblock \emph{arXiv preprint arXiv:2304.10592}, 2023.

\end{thebibliography}
}

\clearpage
\appendix
\setcounter{page}{1}
\maketitlesupplementary

\section{Detailed Experimental Settings}
\label{sec:implementation}

Here we describe the detailed experimental settings to guarantee the reproducibility. All experiments are conducted on NVIDIA A100-80GB GPUs.

\subsection{Datasets}
\label{sec:datasets}

In this work, we adopt 10 datasets in total to validate the effectiveness of the proposed TVP. We categorize them into 4 visual or multimodal tasks and we will introduce them respectively.

\textbf{Object Recognition. }Following~\cite{alayrac2022flamingo,Dai2023InstructBLIP}, we take close-ended evaluation for recognition, restricting the vocabulary to the category names of the datasets. To be specific, The prompt given to the models is ``\textit{This is a photo of a}'' and the target for text completion will be the ground-truth label in text. We concatenate each candidate category after the prompt and select the one with maximum log-likelihood as the prediction. The description for TSE is in the template of ``\textit{This is a photo of a \{\text{ground-truth label}\}}''.

We take 7 datasets for this task, including CIFAR-10, CIFAR-100~\cite{krizhevsky2009learning}, ImageNette~\cite{imagenette} (a subset of ImageNet), which are commonly used for image classification, and SVHN~\cite{svhn}, Oxford Pets~\cite{parkhi12pets}, FGVCAircraft~\cite{maji13fine-grained} (manufacturer level), Food101~\cite{food101}, which are popular datasets for fine-grained classification in specific domains. By default, we take the \texttt{train} split for training, \texttt{val} split for validation and \texttt{test} split for testing as provided in the dataset. If \texttt{val} split is not provided, we sample a certain proportion for validation.

\textbf{Object Counting. }We take CLEVR~\cite{johnson2017clevr} as an example. Unlike recognition, we take an open-ended evaluation for object counting. We ask the models ``\textit{How many objects are there in this image? Answer with a single number.}'' and generate the response with do\_sample set False and other parameters as default. We evaluate the response as correct or not by checking whether the answer of number appears in it. The corresponding description for TSE is ``\textit{There are \{number\} objects in this image}''. We take the \texttt{train} split for training and sample 10\% and 20\% out of \texttt{val} split for validation and testing respectively.

\textbf{Multimodal Reasoning. }We take Hatefulmemes~\cite{kiela2020hateful} for multimodal reasoning, which ask the models to decide whether the text on the meme and the visual content combined together convey hatred. Following~\cite{Dai2023InstructBLIP}, the prompt is ``\textit{This is an image with ``\{\}'' written on it. Is it hateful?}'', and we take the ranking method used for recognition here with ``Yes'' and ``No'' as labels. We use the normalized log-likelihood to calculate ROC AUC score. The description for TSE is ``\textit{This is (not) hateful}''.  We take 90\% of \texttt{train} split for training, the rest 10\% for validation and \texttt{dev} split for testing.

\textbf{Hallucination Correction. }We take POPE~\cite{li2023evaluating}, which ask the models whether there is a certain object in the image or not to evaluate their hallucination. The prompt given to the model is consistent with the default setting in official code, as ``\textit{Is there a ``\{\}'' in the image?}'' and we also take ``Yes'' and ``No'' as labels. The description for TSE is in the template of ``\textit{There are \{object list\} in the image.}'' based on the annotations from MSCOCO~\cite{mscoco}. We take the public release split (3000 samples) for testing and generate another dataset of 12000 samples for training and validation with 90\%-10\% random split. In this work, we only adopt datasets built with adversarial negative sampling strategy to challenge the models at utmost.

\subsection{Models}

We select 6 modern MLLMs for experiments. These models have different implementations, for instance BLIVA~\cite{hu2023bliva} uses two projection layers to better address visual-text alignment and VPGTrans~\cite{zhang2023transfer} introduces the concept of visual prompt generator to transfer pre-trained visual encoder across different LLMs. We clone the official codebase of different models and unify the interface for training and inference to better incorporate different models. 

The detailed configuration for them mainly involves the the choices of LLMs. We take Vicuna-7B-v0~\cite{vicuna2023} for MiniGPT-4~\cite{zhu2023minigpt}, BLIVA and VPGTrans, Vicuna-7B-v1.1 for InstructBLIP~\cite{Dai2023InstructBLIP}, Flan-T5-XL~\cite{weifinetuned} for BLIP2~\cite{li2023blip}, and ChatGLM-6B~\cite{zeng2022glm} for VisualGLM-6B~\cite{visualglm}. For visual encoders, these MLLMs share the structure of ViT-G/14, but with different projection layers and training paradigms, which guarantee the model diversity. These models can be deployed conveniently following the official instructions provided in the repositories. 

As for the CLIP's visual encoder for TSE, we use ViT-B/32, a lightweight and popular version for studying CLIP. Since TSE is to introduce extra task knowledge, it does not need to have the same visual encoder as MLLMs.

\subsection{Hyperparameters}
\label{sec:hyper}

We introduce the setting of hyperparameters in this work. The design of visual prompts has been introduced in~\cref{sec:prelim}. The batch size for training is 16. The learning rate $\gamma$ in~\cref{eq:update} is 10 by default. The maximal number of training epochs is 10 with cosine scheduler following~\cite{wu2022unleashing}.

For the weights for the proposed FCA and TSE loss terms, we set them optimal by searching within \{0.0005, 0.001, 0.003, 0.005, 0.008\} and \{0.0001, 0.0005, 0.001\} respectively on validation set, while keeping other hyperparameters consistent with baselines.

\section{Additional Results}
\label{sec:result}

\subsection{Results on Other Datasets}

Besides the 6 datasets displayed in the main paper, we also validate the effectiveness of our method on 4 commonly used classification datasets and demonstrate the results in~\cref{tab:addition_results}.

Apart from the coarse-grained classification dataset CIFAR-100, the zero-shot performance of modern MLLMs on these fine-grained datasets in specific domains is far from satisfactory, further emphasizing the necessities for adapting MLLMs to downstream tasks. 

The observations and conclusions in~\cref{sec:main_results} remain consistent. We can see that visual prompts generated by TVP on a single model (MiniGPT-4 or InstructBLIP) bring the most significant improvements to 6 models. Moreover, by ensembling two models for training visual prompts, the performance is further boosted to higher levels.

\subsection{Results on Corrupted Datasets}
\label{sec:corruption}

Robustness has been a crucial issue for deep neural network, concerning the stability of model in applications. It is natural to evaluate the robustness of visual prompts to image common corruptions~\cite{hendrycks2019robustness}. We examine the performance of visual prompts generated by MiniGPT-4 on corrupted datasets like CIFAR-10-C and ImageNette-C. We set the severity level as 3 and test with 15 corruptions. We use the official release of CIFAR-10-C and the official code\footnote{https://github.com/hendrycks/robustness} to generate corresponding corrupted dataset for ImageNette.

The results are shown in~\cref{tab:result_corruption}. Visual prompts generated by VP and EVP cannot effectively improve the 6 models on average under the corruptions imposed to CIFAR-10, while TVP can still bring 2.30\% and 3.09\% on CIFAR-10-C and ImageNette-C respectively. The results indicate that the consolidation of task-agnostic representations and enhancement of task-related semantics by TVP effectively strengthen the robustness of learned visual prompts to common image corruptions.

\subsection{Detailed Results for Ablations and Analyses}
\label{sec:detailed_results}

Due to space limit, we only report the average performance or average delta in performance for ablation studies in~\cref{sec:ablation} and in-depth analyses in~\cref{sec:analyses}. Here, we display the results for each setting and each model in detail. Detailed results for~\cref{tab:ablation} are in~\cref{tab:detailed_ablation}, those for~\cref{tab:width} are in~\cref{tab:detailed_width}, those for~\cref{fig:data-scale} are in~\cref{tab:detailed_datascale} and those for~\cref{tab:generalization} are in~\cref{tab:detailed_generalization}.

\section{Discussion on Computational Efficiency}
\label{sec:efficiency}

As we target on efficient adaptation for diverse MLLMs rather than fine-tuning each of them respectively, we here discuss the computational efficiency of the proposed TVP.

\subsection{Comparison with Fine-tuning Methods}

We conduct additional experiments on an A100-80G GPU with half precision and the same batch size as TVP. If the training exceeds GPU memory (e.g., BLIVA), we adopt gradient accumulation. Here we use CIFAR-10 and the prompts trained on InstructBLIP to compare with full fine-tuning and LoRA. Results are displayed in~\cref{tab:comparison}. Though FFT and LoRA have moderately higher accuracy than TVP due to much larger numbers of trainable parameters ($\ge$4B for FFT, $\ge$8M for LoRA and $\sim$70K for TVP), TVP has the minimal computation overhead, which is reflected in the smallest memory demand and the shortest average training time. When the computation resources are limited to fine-tuning, off-the-shelf visual prompts trained by TVP are expected to achieve black-box adaptation with no cost. This supports the motivation of our method.

\begin{table}[h]
    \centering
    \resizebox{\linewidth}{!}{
    \setlength{\tabcolsep}{1.5pt}
    \begin{tabular}{c|ccc|ccc|ccc|ccc}
    \toprule\toprule
          &  \multicolumn{3}{c|}{InstructBLIP}&  \multicolumn{3}{>{\columncolor[gray]{0.6}}c|}{BLIP2} & \multicolumn{3}{>{\columncolor[gray]{0.6}}c|}{MiniGPT-4} & \multicolumn{3}{>{\columncolor[gray]{0.6}}c}{BLIVA}\\\midrule
          &  FFT  & LoRA & TVP   &  FFT  & LoRA & TVP  &  FFT  & LoRA & TVP &  FFT  & LoRA & TVP \\\midrule
         Acc (\%)& 99.16 & 98.78 &98.07 & 99.09 & 98.08 &96.02 &  99.27  &  95.18  & 91.69  & 99.07  & 98.14 &97.78   \\
         Mem. (GB)& 63.5 & 33.8 &31.1 & 36.9 & 21.8 &9.2$^\dag$ &  62.4  &  35.6  &  18.3$^\dag$  &66.5 & 55.2 &18.5$^\dag$  \\
         Time (min)& 30 & 26 &27 & 28 & 26 &0 & 29 & 25 &0 &  118 & 92 &0   \\\bottomrule\bottomrule
    \end{tabular}}
    \caption{Comparison of performance, memory costs and training time with fine-tuning methods. \textcolor{gray}{gray} for black-box models, $\dag$ for inference mode, since they need no training for TVP.}
    \label{tab:comparison}
\end{table}

\subsection{Comparison with Baseline Visual Prompting}

Compared to the baselines, VP and EVP, TVP demands additional forward passes through vision encoders. Taking MiniGPT-4 for example, VP and EVP need one forward pass in each iteration and take around 820GFLOPs. For TVP, the combination of FCA and TSE demands an extra forward pass through the MLLM's visual encoder ($\sim$260GFLOPs, FCA) and another forward pass through CLIP ($\sim$7GFLOPs, TSE). While extra computation for TSE is negligible, FCA brings around 32\% more computation overloads, with a similar increase in training time. However, the original visual features only need to be computed once, thus the cost for FCA can be distributed to each epoch and will only bring around 3\% extra computations when trained for 10 epochs, which is acceptable. The computation overheads can be further alleviated in the future.

\begin{table}  
    \centering
    \begin{subtable}{\linewidth}  
        \centering
        \scriptsize
        \setlength{\tabcolsep}{3pt}
        \resizebox{\linewidth}{!}{
        \begin{tabular}{c|c|cccccc>{\columncolor[gray]{0.9}}c}
        \toprule\toprule
        \multicolumn{2}{c|}{\bf Recognition: CIFAR-100} & \parbox[t]{2mm}{\rotatebox[origin=c]{90}{ MiniGPT-4}}  & \parbox[t]{2mm}{\rotatebox[origin=c]{90}{ InstructBLIP}}  & \parbox[t]{2mm}{\rotatebox[origin=c]{90}{ BLIP2}}  & \parbox[t]{2mm}{\rotatebox[origin=c]{90}{ VPGTrans}}  & \parbox[t]{2mm}{\rotatebox[origin=c]{90}{ BLIVA}}  & \parbox[t]{2mm}{\rotatebox[origin=c]{90}{ VisualGLM}} &  Avg.$\Delta$\\\midrule
                          & Clean & 61.85  & 58.41  &  60.65 & 58.00  & 56.34  & 12.71 & \color{gray}{0.00} \\\midrule
        \multirow{3}{*}{MiniGPT-4} & VP~\cite{bahng2022exploring} & 63.54$^\ast$ & 44.40 & 60.05 & 59.93 & 53.36 & 12.15 & -2.42\\
                          & EVP~\cite{wu2022unleashing}  & 71.05$^\ast$ & 48.91 & 56.43 & 59.23 & 56.44 & 20.10 & +0.70 \\
                          &TVP (ours)   &\bf 75.36$^\ast$  & 65.10 & 64.15 & 57.84 & 53.58 & \bf 21.34 &  \color{purple}\bf +4.90\\\midrule
        \multirow{3}{*}{InstructBLIP} & VP~\cite{bahng2022exploring} & 60.65 & 76.16$^\ast$ & 58.60 & 58.32 & 58.40 & 9.47 & +2.27\\
                          & EVP~\cite{wu2022unleashing}  & 62.24 & \bf 78.68$^\ast$ & 61.66 & 57.37 & 59.86 & 12.13 & +4.00\\
                          &TVP (ours)   & 63.92 & 77.92$^\ast$ & 63.72 & 62.62 & 56.09 & 12.97 & \color{purple}\bf +4.88\\\midrule
        \multirow{3}{*}{Ensemble} & VP~\cite{bahng2022exploring} & 65.48$^\ast$ & 71.77$^\ast$ & 63.48 & 60.25 & 55.04 & 9.15 & +2.87\\
                          & EVP~\cite{wu2022unleashing}  & 70.13$^\ast$ & 74.89$^\ast$ & 62.40 & 62.07 & 60.76 & 14.96 & +6.21\\
                          &TVP (ours)   & 73.33$^\ast$ & 77.62$^\ast$ & \bf 64.19 &\bf 62.79 & \bf 62.18 & 13.26 & \color{purple}\bf +7.57\\\bottomrule
        \end{tabular}}
    \end{subtable}

    \begin{subtable}{\linewidth}  
        \centering
        \scriptsize
        \setlength{\tabcolsep}{3pt}
        \resizebox{\linewidth}{!}{
        \begin{tabular}{c|c|cccccc>{\columncolor[gray]{0.9}}c}
        \toprule
        \multicolumn{2}{c|}{\bf Recognition: Pet37} & \parbox[t]{2mm}{\rotatebox[origin=c]{90}{ MiniGPT-4}}  & \parbox[t]{2mm}{\rotatebox[origin=c]{90}{ InstructBLIP}}  & \parbox[t]{2mm}{\rotatebox[origin=c]{90}{ BLIP2}}  & \parbox[t]{2mm}{\rotatebox[origin=c]{90}{ VPGTrans}}  & \parbox[t]{2mm}{\rotatebox[origin=c]{90}{ BLIVA}}  & \parbox[t]{2mm}{\rotatebox[origin=c]{90}{ VisualGLM}} &  Avg.$\Delta$\\\midrule
                          & Clean & 30.50  &  27.23 & 11.53  & 16.52  & 22.21  & 31.07 & \color{gray}{0.00} \\\midrule
        \multirow{3}{*}{MiniGPT-4} & VP~\cite{bahng2022exploring} & 42.38$^\ast$ & 33.69 & 11.80 & 23.14 & 25.81 & 29.46 & +4.54\\
                          & EVP~\cite{wu2022unleashing}  & 56.67$^\ast$ & 30.44 & 13.22 & 22.40 & 27.91 & 28.37 & +6.66 \\
                          &TVP (ours)   &\bf 59.53$^\ast$  & 39.00 & \bf 16.57 & 25.27 & 30.53 & 29.35 &\color{purple}\bf +10.20 \\\midrule
        \multirow{3}{*}{InstructBLIP} & VP~\cite{bahng2022exploring} & 40.23 & 37.80$^\ast$ & 13.27 & 17.83 & 29.71 & 29.54 & +4.89\\
                          & EVP~\cite{wu2022unleashing}  & 40.83 & 65.25$^\ast$ & 12.16 & 17.63 & 31.70 & 30.36 & +9.81 \\
                          &TVP (ours)   & 41.05 & \bf 66.86$^\ast$ & 14.28 & 22.27 & 42.95 & 30.44 & \color{purple}\bf +13.13 \\\midrule
        \multirow{3}{*}{Ensemble} & VP~\cite{bahng2022exploring} & 46.77$^\ast$ & 43.80$^\ast$ & 15.10 & 22.13 & 32.73 & \bf 30.69 & +8.69\\
                          & EVP~\cite{wu2022unleashing}  & 56.99$^\ast$ & 66.31$^\ast$ & 13.55 & 15.10 & 32.76 & 29.60 & +12.54\\
                          &TVP (ours)   & 51.35$^\ast$ & 61.60$^\ast$ & 13.87 & \bf 26.30 & \bf48.11 & 28.02 &  \color{purple}\bf +15.03\\\bottomrule
        \end{tabular}}
    \end{subtable}

    \begin{subtable}{\linewidth}  
        \centering
        \scriptsize
        \setlength{\tabcolsep}{3pt}
        \resizebox{\linewidth}{!}{
        \begin{tabular}{c|c|cccccc>{\columncolor[gray]{0.9}}c}
        \toprule
        \multicolumn{2}{c|}{\bf Recognition: Aircraft} & \parbox[t]{2mm}{\rotatebox[origin=c]{90}{ MiniGPT-4}}  & \parbox[t]{2mm}{\rotatebox[origin=c]{90}{ InstructBLIP}}  & \parbox[t]{2mm}{\rotatebox[origin=c]{90}{ BLIP2}}  & \parbox[t]{2mm}{\rotatebox[origin=c]{90}{ VPGTrans}}  & \parbox[t]{2mm}{\rotatebox[origin=c]{90}{ BLIVA}}  & \parbox[t]{2mm}{\rotatebox[origin=c]{90}{ VisualGLM}} &  Avg.$\Delta$\\\midrule
                          & Clean & 8.55  & 10.26  & 6.54  &  14.34 & 8.19  & 4.05 & \color{gray}{0.00} \\\midrule
        \multirow{3}{*}{MiniGPT-4} & VP~\cite{bahng2022exploring} & 30.15$^\ast$ & 8.67 & 6.93 & 14.97 & 11.13 & 4.02 & +3.99\\
                          & EVP~\cite{wu2022unleashing}  & 32.52$^\ast$ & 9.36 & 6.42 & 17.64 & 11.25 & 4.02 & +4.88 \\
                          &TVP (ours)   & \bf 33.99$^\ast$  & 9.81 &\bf 7.41 & 20.76 & 7.20 & 4.02 & \color{purple}\bf +5.21\\\midrule
        \multirow{3}{*}{InstructBLIP} & VP~\cite{bahng2022exploring} & 12.90 & 16.92$^\ast$ & 4.59 & 12.18 & 8.97 & 4.02 & +1.27\\
                          & EVP~\cite{wu2022unleashing}  & 22.92 & 31.35$^\ast$ & 5.28 & \bf 23.97 & 11.04 & 4.02 & +7.78\\
                          &TVP (ours)   & 30.48 & \bf 36.03$^\ast$ & 4.02 & 20.76 & 11.85 & 4.04 & \color{purple}\bf  +9.21\\\midrule
        \multirow{3}{*}{Ensemble} & VP~\cite{bahng2022exploring} & 28.68$^\ast$ & 25.50$^\ast$ & 7.29 & 13.02 & 11.52 & 4.05 & +6.35\\
                          & EVP~\cite{wu2022unleashing}  & 26.76$^\ast$ & 26.34$^\ast$ & 6.45 & 17.13 & 12.22 & 4.02 & +6.83\\
                          &TVP (ours)   & 30.27$^\ast$ & 24.84$^\ast$ & 4.02 & 23.10 & \bf 24.00 & \bf 4.59 & \color{purple}\bf  +9.82\\\bottomrule
        \end{tabular}}
    \end{subtable}

    \begin{subtable}{\linewidth}  
        \centering
        \scriptsize
        \setlength{\tabcolsep}{3pt}
        \resizebox{\linewidth}{!}{
        \begin{tabular}{c|c|cccccc>{\columncolor[gray]{0.9}}c}
        \toprule
        \multicolumn{2}{c|}{\bf Recognition: Food101} & \parbox[t]{2mm}{\rotatebox[origin=c]{90}{ MiniGPT-4}}  & \parbox[t]{2mm}{\rotatebox[origin=c]{90}{ InstructBLIP}}  & \parbox[t]{2mm}{\rotatebox[origin=c]{90}{ BLIP2}}  & \parbox[t]{2mm}{\rotatebox[origin=c]{90}{ VPGTrans}}  & \parbox[t]{2mm}{\rotatebox[origin=c]{90}{ BLIVA}}  & \parbox[t]{2mm}{\rotatebox[origin=c]{90}{ VisualGLM}} &  Avg.$\Delta$\\\midrule
                          & Clean & 32.99  & 28.99  &  47.29 & 30.42  & 36.08  & 5.90 & \color{gray}{0.00} \\\midrule
        \multirow{3}{*}{MiniGPT-4} & VP~\cite{bahng2022exploring} & 50.14$^\ast$ & 32.08  & 34.10  & 23.49  & 31.92  & 4.67  & -0.88 \\
                          & EVP~\cite{wu2022unleashing}  & 63.72$^\ast$ & 30.93 & 45.43 & 27.64 & 33.74 & 3.88 & +3.95 \\
                          &TVP (ours)   & \bf 64.16$^\ast$  & 37.66 & 48.95 & 29.43 & 36.36 & 5.54 & \color{purple}\bf  +6.74 \\\midrule
        \multirow{3}{*}{InstructBLIP} & VP~\cite{bahng2022exploring} & 19.68 & 41.23$^\ast$  & 33.70  & 26.85  & 36.28  & \bf 8.71  & -2.54 \\
                          & EVP~\cite{wu2022unleashing}  & 37.47 & 64.95$^\ast$ & 48.87 & 31.25 & 43.37 & 3.84 & +8.01\\
                          &TVP (ours)   & 38.49  & \bf 68.51$^\ast$  &  48.55 & 31.13  &  44.75  & 6.02  & \color{purple}\bf  +9.46 \\\midrule
        \multirow{3}{*}{Ensemble} & VP~\cite{bahng2022exploring} & 53.03$^\ast$ & 59.21$^\ast$  & 47.29 & 27.68  & \bf 46.57  & 6.42  & +9.75 \\
                          & EVP~\cite{wu2022unleashing}  & 63.48$^\ast$ & 66.50$^\ast$ & 48.20 & 27.49 & 26.46 & 4.12 & +9.10\\
                          &TVP (ours)   & 63.92$^\ast$ & 66.22$^\ast$ & \bf 51.80 & \bf 34.61 & 44.44 & 5.47 & \color{purple}\bf  +14.13 \\\bottomrule\bottomrule
        \end{tabular}}
    \end{subtable}
    \caption{Results on 4 more datasets of object recognition. Visual prompts are trained on MiniGPT-4, InstructBLIP and their ensemble with different methods, and further tested on 6 modern MLLMs. Top-1 accuracy (\%) is reported.}
\label{tab:addition_results}
\end{table}

\begin{table}[]
\centering
\setlength{\tabcolsep}{3pt}
\resizebox{\linewidth}{!}{%
\begin{tabular}{l|c|cccccc|c}
\toprule
Trained on                    & \parbox[t]{2mm}{\rotatebox[origin=c]{90}{Prompt Wid.}} &\parbox[t]{2mm}{\rotatebox[origin=c]{90}{MiniGPT-4}}  & \parbox[t]{2mm}{\rotatebox[origin=c]{90}{ InstructBLIP}}  & \parbox[t]{2mm}{\rotatebox[origin=c]{90}{ BLIP2}}  & \parbox[t]{2mm}{\rotatebox[origin=c]{90}{ VPGTrans}}  & \parbox[t]{2mm}{\rotatebox[origin=c]{90}{ BLIVA}}  & \parbox[t]{2mm}{\rotatebox[origin=c]{90}{ VisualGLM}}   & Avg. $\Delta$ \\\midrule
\multirow{6}{*}{MiniGPT-4}    & 5            & 94.29 & 84.17 & 89.40 & 91.60 & 90.94 & 90.15 &   +2.84            \\
                              & 10           & 96.00 & 84.03 & 93.17 & 91.82 & 92.73 & 85.17 &     +3.23          \\
                              & 20           & 96.82 & 91.26 & 86.68 & 88.49 & 93.71 & 90.39 &     +3.97          \\
                              & 40           & 95.70 & 89.44 & 88.49 & 87.68 & 89.69 & 88.78 &      +2.71         \\
                              & 50           & 96.77 & 87.29 & 87.62 & 88.26 & 88.05 & 87.15 &     +1.93          \\
                              & 80           & 94.21 & 78.21 & 86.41 & 84.02 & 85.91 & 76.62 &      -3.03         \\\midrule
\multirow{6}{*}{InstructBLIP} & 5            & 89.73 & 96.41 & 85.08 & 91.95 & 93.64 & 88.97 &   +3.71            \\
                              & 10           & 88.03 & 97.06 & 92.47 & 91.89 & 94.90 & 91.72 &     +5.42          \\
                              & 20           & 88.04 & 98.04 & 82.78 & 93.96 & 97.95 & 93.13 &    +5.06           \\
                              & 40           & 85.16 & 98.24 & 86.37 & 89.21 & 89.88 & 93.55 &    +3.15           \\
                              & 50           & 84.52 & 97.81 & 93.58 & 87.33 & 91.50 & 86.28 &    +2.91           \\
                              & 80           & 82.38 & 94.75 & 83.15 & 81.85 & 88.31 & 80.39 &     -2.12          \\\midrule
\multirow{6}{*}{Ensemble}     & 5            & 91.64 & 94.53 & 94.72 & 88.97 & 94.71 & 86.94 &     +4.66          \\
                              & 10           & 95.08 & 95.65 & 92.73 & 87.58 & 94.00 & 78.77 &      +3.38         \\
                              & 20           & 95.19 & 96.55 & 93.37 & 90.59 & 96.23 & 84.33 &      +5.45         \\
                              & 40           & 97.73 & 97.41 & 84.59 & 91.52 & 97.17 & 90.34 &      +5.87         \\
                              & 50           & 96.60 & 97.31 & 88.85 & 88.20 & 95.02 & 88.31 &      +5.12         \\
                              & 80           & 92.01 & 95.99 & 84.14 & 81.64 & 94.14 & 91.62 &  +2.67  \\\bottomrule          
\end{tabular}}
\caption{Detaile results for the ablation study about the impact of prompt width on the performance of TVP on CIFAR-10 in~\cref{tab:width}.}
\label{tab:detailed_width}
\end{table}

\begin{table}[]
\centering
\setlength{\tabcolsep}{2pt}
\resizebox{\linewidth}{!}{%
\begin{tabular}{c|l|cccccc|c}
\toprule
Datasets & Model & \parbox[t]{2mm}{\rotatebox[origin=c]{90}{MiniGPT-4}}  & \parbox[t]{2mm}{\rotatebox[origin=c]{90}{ InstructBLIP}}  & \parbox[t]{2mm}{\rotatebox[origin=c]{90}{ BLIP2}}  & \parbox[t]{2mm}{\rotatebox[origin=c]{90}{ VPGTrans}}  & \parbox[t]{2mm}{\rotatebox[origin=c]{90}{ BLIVA}}  & \parbox[t]{2mm}{\rotatebox[origin=c]{90}{ VisualGLM}} & Avg. $\Delta$ \\\midrule
\multirow{3}{*}{CIFAR-10} & VP~\cite{bahng2022exploring} & 87.97 & 94.20 & 82.59 & 88.64 & 89.13 & 90.47 & +1.57 \\
&EVP~\cite{wu2022unleashing} & 81.89 & 89.44 & 82.19 & 90.11 & 84.15 & 95.14 & -0.11 \\
&TVP (ours) & 88.80 & 94.38 & 89.92 & 91.53 & 89.91 & 93.63 & +4.10 \\\midrule
\multirow{3}{*}{ImageNette}& VP~\cite{bahng2022exploring} & 84.25 & 74.70 & 92.00 & 81.50 & 84.59 & 72.79 & +1.24 \\
&EVP~\cite{wu2022unleashing} & 83.18 & 77.38 & 87.39 & 83.31 & 79.82 & 72.18 & +0.14 \\
&TVP (ours) & 88.36 & 77.20 & 94.01 & 82.78 & 80.15 & 73.86 & +2.33 \\\midrule
\multirow{3}{*}{SVHN}&VP~\cite{bahng2022exploring} & 39.42 & 30.00 & 32.87 & 33.57 & 27.49 & 21.62 & -0.29 \\
&EVP~\cite{wu2022unleashing} & 35.34 & 24.27 & 33.20 & 34.63 & 21.97 & 20.24 & -2.84 \\
&TVP (ours) & 41.98 & 30.02 & 26.88 & 39.49 & 31.55 & 26.23 & +1.57 \\\midrule
\multirow{3}{*}{Pet37}&VP~\cite{bahng2022exploring} & 34.15 & 33.01 & 14.64 & 20.82 & 25.40 & 28.26 & +2.87 \\
&EVP~\cite{wu2022unleashing} & 33.39 & 31.34 & 9.46 & 16.54 & 29.35 & 26.25 & +1.21 \\
&TVP (ours) & 38.05 & 30.01 & 14.99 & 23.58 & 28.56 & 27.12 & +3.87 \\\midrule
\multirow{3}{*}{Aircraft}&VP~\cite{bahng2022exploring} & 16.50 & 7.74 & 5.88 & 13.08 & 10.56 & 4.02 & +0.97 \\
&EVP~\cite{wu2022unleashing} & 19.05 & 9.24 & 4.05 & 11.01 & 9.09 & 4.02 & +0.75 \\
&TVP (ours) & 15.15 & 11.58 & 4.05 & 10.68 & 8.58 & 4.02 & +0.35 \\\midrule
\multirow{3}{*}{Food101}&VP~\cite{bahng2022exploring} & 31.45 & 27.33 & 40.48 & 29.19 & 41.70 & 5.31 & -1.04 \\
&EVP~\cite{wu2022unleashing} & 33.03 & 33.27 & 37.90 & 26.85 & 40.24 & 4.08 & -1.05 \\
&TVP (ours) & 37.47 & 43.84 & 38.89 & 28.95 & 38.93 & 4.95 & +1.89 \\\bottomrule
\end{tabular}%
}
\caption{Detailed results for the analysis on the generalization of TVP using ensemble across diverse recognition datasets in~\cref{tab:generalization}.}
\label{tab:detailed_generalization}
\end{table}

\renewcommand\theadfont{\scriptsize}
\begin{table*}
    \centering
    \begin{subtable}{\linewidth}
        \centering
        \scriptsize
        \setlength{\tabcolsep}{1pt}
        \resizebox{\linewidth}{!}{
            \begin{tabular}{l|ccccccccccccccc>{\columncolor[gray]{0.9}}c}
            \toprule\toprule
\thead{Corruption\\Types} & Fog   & \thead{JPEG\\Compression} & \thead{Zoom\\Blur} & \thead{Glass\\Blur} & \thead{Shot\\Noise} & \thead{Defocus\\Blur} & \thead{Elastic\\Transform} & Frost & Brightness & Snow  & \thead{Gaussian\\noise} & \thead{Motion\\Blur} & Contrast & \thead{Impulse\\Noise} & Pixelate &   Avg.$\Delta$    \\\midrule
Clean      & 85.73 & 69.41            & 82.87     & 70.72      & 71.13      & 85.93        & 82.87             & 83.12 & 86.69      & 82.69 & 65.85          & 80.24       & 86.72    & 79.38         & 81.75    & \color{gray}{0.00}  \\
VP~\cite{bahng2022exploring}         & 81.96 & 55.36            & 79.68     & 58.69      & 60.05      & 82.67        & 80.42             & 77.66 & 83.59      & 78.71 & 52.67          & 74.12       & 82.86    & 71.06         & 76.26    & -6.62 \\
EVP~\cite{wu2022unleashing}        & 85.31 & 57.10            & 82.51     & 58.49      & 63.76      & 85.93        & 83.90             & 80.46 & 86.78      & 82.33 & 56.20          & 76.57       & 85.96    & 75.04         & 76.67    & -3.87 \\
TVP (ours)        & 89.46 & 68.65            & 87.46     & 67.95      & 71.92      & 89.95        & 87.85             & 85.58 & 90.82      & 86.76 & 66.12          & 83.01       & 90.02    & 80.71         & 83.34    &\color{purple}\bf +2.30  \\\bottomrule\bottomrule
\end{tabular}%
        }
    \caption{Average performance under different common corruptions at level 3 on CIFAR-10 with visual prompts generated on MiniGPT-4.}
    \end{subtable}

    \begin{subtable}{\linewidth}
        \centering
        \scriptsize
        \setlength{\tabcolsep}{1pt}
        \resizebox{\linewidth}{!}{
            \begin{tabular}{l|ccccccccccccccc>{\columncolor[gray]{0.9}}c}
            \toprule\toprule
\thead{Corruption\\Types} & Fog   & \thead{JPEG\\Compression} & \thead{Zoom\\Blur} & \thead{Glass\\Blur} & \thead{Shot\\Noise} & \thead{Defocus\\Blur} & \thead{Elastic\\Transform} & Frost & Brightness & Snow  & \thead{Gaussian\\noise} & \thead{Motion\\Blur} & Contrast & \thead{Impulse\\Noise} & Pixelate &   Avg.$\Delta$    \\\midrule
Clean      & 79.15 & 80.50            & 69.91     & 71.26      & 76.77      & 75.92        & 72.24             & 74.36 & 79.71      & 76.23 & 76.96          & 76.52       & 79.44    & 76.98         & 81.15    & \color{gray}{0.00} \\
VP~\cite{bahng2022exploring}         & 78.76 & 80.62            & 69.40     & 71.77      & 78.09      & 76.36        & 72.82             & 73.78 & 80.72      & 75.57 & 78.31          & 76.74       & 78.92    & 78.42         & 80.54    & +0.25 \\
EVP~\cite{wu2022unleashing}        & 79.33 & 81.58            & 67.01     & 70.95      & 78.52      & 76.74        & 73.44             & 73.38 & 82.07      & 75.68 & 78.76          & 76.78       & 78.46    & 78.65         & 82.18    & +0.43 \\
TVP (ours)       & 82.53 & 84.20            & 70.31     & 73.25      & 81.00      & 80.32        & 74.76             & 76.51 & 83.58      & 79.19 & 81.07          & 79.85       & 82.07    & 81.05         & 83.79    & \color{purple}\bf+3.09\\\bottomrule\bottomrule
\end{tabular}%
        }
    \caption{Average performance under different common corruptions at level 3 on ImageNette with visual prompts generated on MiniGPT-4.}
    \end{subtable}
\vspace{-3ex}
\caption{Average performance under common corruptions~\cite{hendrycks2019robustness} of different methods on CIFAR-10 and ImageNette. Visual prompts generated by the proposed TVP still lead to the most significant improvements, showing better robustness to common corruptions.}
\label{tab:result_corruption}
\end{table*}

\begin{table*}[]
    \centering
    \begin{subtable}{.495\linewidth}
        \centering
            \scriptsize
            \setlength{\tabcolsep}{3pt}
            \resizebox{\linewidth}{!}{
            \begin{tabular}{cc|ccccccc}
            \toprule
            FCA & TSE & \parbox[t]{2mm}{\rotatebox[origin=c]{90}{ MiniGPT-4}}  & \parbox[t]{2mm}{\rotatebox[origin=c]{90}{ InstructBLIP}}  & \parbox[t]{2mm}{\rotatebox[origin=c]{90}{ BLIP2}}  & \parbox[t]{2mm}{\rotatebox[origin=c]{90}{ VPGTrans}}  & \parbox[t]{2mm}{\rotatebox[origin=c]{90}{ BLIVA}}  & \parbox[t]{2mm}{\rotatebox[origin=c]{90}{ VisualGLM}} &  Avg.$\Delta$\\\midrule
           \XSolidBrush  & \XSolidBrush & 97.97 & 84.57 & 83.39 & 86.93 & 86.45 & 85.92 & 0.28 \\
           \Checkmark  & \XSolidBrush & 97.95 & 86.97 & 90.58 & 90.94 & 92.18 & 87.82 & 3.82 \\
           \XSolidBrush  & \Checkmark & 97.94 & 86.93 & 85.74 & 90.32 & 92.78 & 81.30 & 1.91 \\\midrule
           \Checkmark  & \Checkmark & 98.33 & 92.82 & 91.68 & 88.70 & 87.48 & 87.53 & 3.83  \\\bottomrule
        \end{tabular}}
        \caption{CIFAR-10}
    \end{subtable}
\hfill
    \begin{subtable}{.495\linewidth}
        \centering
            \scriptsize
            \setlength{\tabcolsep}{3pt}
            \resizebox{\linewidth}{!}{
            \begin{tabular}{cc|ccccccc}
            \toprule
            FCA & TSE & \parbox[t]{2mm}{\rotatebox[origin=c]{90}{ MiniGPT-4}}  & \parbox[t]{2mm}{\rotatebox[origin=c]{90}{ InstructBLIP}}  & \parbox[t]{2mm}{\rotatebox[origin=c]{90}{ BLIP2}}  & \parbox[t]{2mm}{\rotatebox[origin=c]{90}{ VPGTrans}}  & \parbox[t]{2mm}{\rotatebox[origin=c]{90}{ BLIVA}}  & \parbox[t]{2mm}{\rotatebox[origin=c]{90}{ VisualGLM}} &  Avg.$\Delta$\\\midrule
           \XSolidBrush & \XSolidBrush & 96.79 & 68.15 & 91.36 & 79.08 & 75.82 & 76.05 & 0.81 \\
\Checkmark   & \XSolidBrush & 95.87 & 75.87 & 96.51 & 82.24 & 78.19 & 70.11 & 2.73 \\
\XSolidBrush & \Checkmark   & 97.81 & 67.59 & 91.64 & 78.35 & 84.23 & 82.93 & 3.36 \\ \midrule
\Checkmark   & \Checkmark   & 97.71 & 78.34 & 94.98 & 86.34 & 84.51 & 75.34 & 5.80 \\\bottomrule
        \end{tabular}}
        \caption{ImageNette}
    \end{subtable}

    \begin{subtable}{.495\linewidth}
        \centering
            \scriptsize
            \setlength{\tabcolsep}{3pt}
            \resizebox{\linewidth}{!}{
            \begin{tabular}{cc|ccccccc}
            \toprule
            FCA & TSE & \parbox[t]{2mm}{\rotatebox[origin=c]{90}{ MiniGPT-4}}  & \parbox[t]{2mm}{\rotatebox[origin=c]{90}{ InstructBLIP}}  & \parbox[t]{2mm}{\rotatebox[origin=c]{90}{ BLIP2}}  & \parbox[t]{2mm}{\rotatebox[origin=c]{90}{ VPGTrans}}  & \parbox[t]{2mm}{\rotatebox[origin=c]{90}{ BLIVA}}  & \parbox[t]{2mm}{\rotatebox[origin=c]{90}{ VisualGLM}} &  Avg.$\Delta$\\\midrule
           \XSolidBrush & \XSolidBrush & 74.24 & 41.59 & 48.87 & 57.61 & 36.11 & 33.12 & 17.48 \\
\Checkmark   & \XSolidBrush & 74.81 & 56.69 & 52.98 & 51.96 & 50.35 & 24.93 & 20.84 \\
\XSolidBrush & \Checkmark   & 81.39 & 53.41 & 50.99 & 59.46 & 56.60  & 32.91 & 24.68 \\\midrule
\Checkmark   & \Checkmark   & 75.17 & 54.32 & 61.95 & 51.10  & 60.28 & 32.17 & 24.72 \\\bottomrule
        \end{tabular}}
        \caption{SVHN}
        
    \end{subtable}
\hfill
    \begin{subtable}{.495\linewidth}
        \centering
            \scriptsize
            \setlength{\tabcolsep}{3pt}
            \resizebox{\linewidth}{!}{
            \begin{tabular}{cc|ccccccc}
            \toprule
            FCA & TSE & \parbox[t]{2mm}{\rotatebox[origin=c]{90}{ MiniGPT-4}}  & \parbox[t]{2mm}{\rotatebox[origin=c]{90}{ InstructBLIP}}  & \parbox[t]{2mm}{\rotatebox[origin=c]{90}{ BLIP2}}  & \parbox[t]{2mm}{\rotatebox[origin=c]{90}{ VPGTrans}}  & \parbox[t]{2mm}{\rotatebox[origin=c]{90}{ BLIVA}}  & \parbox[t]{2mm}{\rotatebox[origin=c]{90}{ VisualGLM}} &  Avg.$\Delta$\\\midrule
           \XSolidBrush & \XSolidBrush & 52.17 & 39.03 & 20.17 & 8.00  & 34.23 & 13.60 & 8.64  \\
\Checkmark   & \XSolidBrush & 50.57 & 36.03 & 22.30 & 20.93 & 32.53 & 18.03 & 10.83 \\
\XSolidBrush & \Checkmark   & 54.07 & 31.33 & 16.60 & 20.33 & 32.87 & 21.07 & 10.15 \\\midrule
\Checkmark   & \Checkmark   & 51.00 & 42.90 & 22.07 & 19.50 & 36.00 & 13.00 & 11.51\\\bottomrule
        \end{tabular}}
        \caption{CLEVR}
        
    \end{subtable}

    \begin{subtable}{.495\linewidth}
        \centering
            \scriptsize
            \setlength{\tabcolsep}{3pt}
            \resizebox{\linewidth}{!}{
            \begin{tabular}{cc|ccccccc}
            \toprule
            FCA & TSE & \parbox[t]{2mm}{\rotatebox[origin=c]{90}{ MiniGPT-4}}  & \parbox[t]{2mm}{\rotatebox[origin=c]{90}{ InstructBLIP}}  & \parbox[t]{2mm}{\rotatebox[origin=c]{90}{ BLIP2}}  & \parbox[t]{2mm}{\rotatebox[origin=c]{90}{ VPGTrans}}  & \parbox[t]{2mm}{\rotatebox[origin=c]{90}{ BLIVA}}  & \parbox[t]{2mm}{\rotatebox[origin=c]{90}{ VisualGLM}} &  Avg.$\Delta$\\\midrule
           \XSolidBrush & \XSolidBrush & 57.58 & 60.66 & 55.34 & 56.87 & 60.64 & 57.27 & 0.09 \\
\Checkmark   & \XSolidBrush & 56.99 & 63.24 & 54.15 & 58.82 & 63.00 & 57.52 & 0.99 \\
\XSolidBrush & \Checkmark   & 58.31 & 61.65 & 55.20 & 56.43 & 61.66 & 56.40 & 0.31 \\\midrule
\Checkmark   & \Checkmark   & 56.93 & 62.38 & 56.20 & 60.19 & 64.09 & 58.15 & 1.69\\\bottomrule
        \end{tabular}}
        \caption{Hatefulmemes}
        
    \end{subtable}
\hfill
    \begin{subtable}{.495\linewidth}
        \centering
            \scriptsize
            \setlength{\tabcolsep}{3pt}
            \resizebox{\linewidth}{!}{
            \begin{tabular}{cc|ccccccc}
            \toprule
            FCA & TSE & \parbox[t]{2mm}{\rotatebox[origin=c]{90}{ MiniGPT-4}}  & \parbox[t]{2mm}{\rotatebox[origin=c]{90}{ InstructBLIP}}  & \parbox[t]{2mm}{\rotatebox[origin=c]{90}{ BLIP2}}  & \parbox[t]{2mm}{\rotatebox[origin=c]{90}{ VPGTrans}}  & \parbox[t]{2mm}{\rotatebox[origin=c]{90}{ BLIVA}}  & \parbox[t]{2mm}{\rotatebox[origin=c]{90}{ VisualGLM}} &  Avg.$\Delta$\\\midrule
           \XSolidBrush & \XSolidBrush & 68.06 & 69.80 & 50.00 & 61.07 & 71.33 & 69.40 & 0.90 \\
\Checkmark   & \XSolidBrush & 69.60 & 74.00 & 50.13 & 59.47 & 74.80 & 70.27 & 2.33 \\
\XSolidBrush & \Checkmark   & 69.00 & 75.13 & 49.93 & 61.27 & 72.40 & 69.47 & 2.15 \\\midrule
\Checkmark   & \Checkmark   & 68.73 & 75.13 & 51.40 & 64.47 & 72.67 & 71.00 & 3.19\\\bottomrule
        \end{tabular}}
        \caption{POPE}
        
    \end{subtable}
    \caption{Detailed results for the ablation study on different combinations of FCA and TSE in~\cref{tab:ablation}.}
    \label{tab:detailed_ablation}
\end{table*}

\begin{table*}
    \begin{subtable}{.495\linewidth}
         \centering
            \scriptsize
            \setlength{\tabcolsep}{3pt}
            \resizebox{\linewidth}{!}{
            \begin{tabular}{l|l|cccccc|c}
            \toprule
            \multicolumn{2}{c|}{Model} & \parbox[t]{2mm}{\rotatebox[origin=c]{90}{MiniGPT-4}}  & \parbox[t]{2mm}{\rotatebox[origin=c]{90}{ InstructBLIP}}  & \parbox[t]{2mm}{\rotatebox[origin=c]{90}{ BLIP2}}  & \parbox[t]{2mm}{\rotatebox[origin=c]{90}{ VPGTrans}}  & \parbox[t]{2mm}{\rotatebox[origin=c]{90}{ BLIVA}}  & \parbox[t]{2mm}{\rotatebox[origin=c]{90}{ VisualGLM}} &  Avg.\\\midrule
            \multicolumn{9}{c}{\bf CIFAR-10}\\ \midrule
            \multirow{5}{*}{VP~\cite{bahng2022exploring}} & 1\% & 83.08 & 74.95 & 79.80 & 80.35 & 81.67 & 75.28 & 79.19 \\
 & 5\% & 82.06 & 76.07 & 79.20 & 80.97 & 80.07 & 77.98 & 79.39 \\
 & 10\% & 84.29 & 77.58 & 80.00 & 80.99 & 81.96 & 73.38 & 79.70 \\
 & 25\% & 90.97 & 80.35 & 82.42 & 81.34 & 84.63 & 77.28 & 82.83 \\
 & 50\% & 90.53 & 81.14 & 77.45 & 83.29 & 84.31 & 79.84 & 82.76 \\\midrule
\multirow{5}{*}{EVP~\cite{wu2022unleashing}} & 1\% & 97.11 & 86.67 & 83.74 & 87.06 & 89.23 & 82.60 & 87.74 \\
 & 5\% & 97.85 & 85.56 & 83.06 & 86.64 & 87.92 & 86.49 & 87.92 \\
 & 10\% & 97.93 & 83.15 & 83.00 & 88.81 & 84.78 & 86.49 & 87.36 \\
 & 25\% & 98.24 & 85.16 & 82.29 & 87.53 & 85.71 & 85.72 & 87.44 \\
 & 50\% & 98.00 & 84.07 & 83.86 & 87.39 & 86.66 & 86.01 & 87.67 \\\midrule
\multirow{5}{*}{TVP (ours)} & 1\% & 97.80 & 86.04 & 85.27 & 87.90 & 88.65 & 89.69 & 89.23 \\
 & 5\% & 97.24 & 87.79 & 88.32 & 87.28 & 90.09 & 89.14 & 89.98 \\
 & 10\% & 97.85 & 87.69 & 90.82 & 87.36 & 91.97 & 87.87 & 90.59 \\
 & 25\% & 98.23 & 84.86 & 89.20 & 87.61 & 89.96 & 86.34 & 89.37 \\
 & 50\% & 97.68 & 87.59 & 93.57 & 86.33 & 88.12 & 85.25 & 89.76 \\\midrule
        \multicolumn{9}{c}{\bf SVHN} \\ \midrule
        \multirow{5}{*}{VP~\cite{bahng2022exploring}} & 1\% & 67.63 & 47.78 & 43.42 & 36.19 & 38.35 & 26.21 & 43.26 \\
 & 5\% & 66.96 & 38.52 & 47.80 & 44.07 & 33.59 & 34.96 & 44.32 \\
 & 10\% & 81.06 & 35.09 & 21.37 & 40.87 & 32.99 & 20.47 & 38.64 \\
 & 25\% & 58.26 & 41.86 & 50.85 & 58.74 & 35.33 & 29.64 & 45.78 \\
 & 50\% & 73.98 & 50.65 & 46.92 & 45.37 & 43.52 & 20.82 & 46.88 \\\midrule
\multirow{5}{*}{EVP~\cite{wu2022unleashing}} & 1\% & 75.05 & 31.80 & 52.44 & 42.97 & 44.37 & 27.53 & 45.69 \\
 & 5\% & 77.55 & 44.95 & 48.97 & 57.99 & 39.28 & 36.55 & 50.88 \\
 & 10\% & 80.57 & 42.22 & 61.53 & 53.69 & 53.69 & 22.24 & 52.32 \\
 & 25\% & 76.78 & 44.75 & 51.18 & 59.47 & 46.58 & 33.14 & 51.98 \\
 & 50\% & 75.98 & 41.35 & 47.41 & 55.47 & 34.93 & 30.37 & 47.59 \\\midrule
\multirow{5}{*}{TVP (ours)} & 1\% & 79.68 & 38.39 & 56.55 & 46.34 & 48.16 & 33.36 & 50.41 \\
 & 5\% & 85.53 & 48.70 & 64.67 & 58.15 & 48.28 & 39.79 & 57.52 \\
 & 10\% & 80.52 & 47.70 & 61.37 & 56.63 & 45.20 & 35.24 & 54.44 \\
 & 25\% & 80.70 & 51.20 & 62.89 & 59.59 & 59.93 & 35.20 & 58.25 \\
 & 50\% & 82.58 & 51.23 & 65.77 & 59.79 & 53.57 & 35.91 & 58.14 \\\midrule
        \multicolumn{9}{c}{\bf Hatefulmemes} \\ \midrule
        \multirow{5}{*}{VP~\cite{bahng2022exploring}} & 1\% & 57.04 & 56.22 & 60.26 & 51.79 & 49.67 & 56.58 & 55.26 \\
 & 5\% & 57.08 & 56.44 & 54.53 & 53.84 & 54.68 & 54.31 & 55.15 \\
 & 10\% & 57.50 & 55.73 & 57.94 & 55.40 & 53.38 & 48.88 & 54.81 \\
 & 25\% & 60.16 & 58.05 & 57.49 & 55.54 & 53.21 & 48.01 & 55.41 \\
 & 50\% & 57.27 & 55.97 & 53.82 & 54.38 & 54.70 & 55.34 & 55.25 \\\midrule
\multirow{5}{*}{EVP~\cite{wu2022unleashing}} & 1\% & 48.33 & 60.13 & 55.84 & 61.23 & 59.54 & 56.94 & 57.00 \\
 & 5\% & 54.29 & 62.50 & 52.35 & 55.90 & 61.70 & 57.02 & 57.29 \\
 & 10\% & 55.02 & 62.45 & 50.84 & 57.80 & 63.04 & 57.31 & 57.74 \\
 & 25\% & 58.26 & 59.88 & 54.90 & 54.60 & 62.16 & 58.92 & 58.12 \\
 & 50\% & 57.42 & 61.64 & 57.07 & 52.98 & 61.64 & 57.64 & 58.07 \\\midrule
\multirow{5}{*}{TVP (ours)} & 1\% & 51.60 & 61.32 & 55.06 & 59.24 & 61.36 & 57.80 & 57.73 \\
 & 5\% & 53.65 & 62.68 & 53.30 & 58.00 & 62.48 & 58.22 & 58.06 \\
 & 10\% & 54.75 & 61.40 & 54.52 & 59.10 & 62.53 & 58.02 & 58.39 \\
 & 25\% & 59.83 & 61.81 & 53.86 & 57.17 & 62.88 & 57.45 & 58.83 \\
 & 50\% & 55.34 & 62.45 & 53.51 & 59.42 & 64.65 & 58.06 & 58.91  \\\bottomrule
        \end{tabular}}
    \end{subtable}
    \hfill
    \begin{subtable}{.495\linewidth}
         \centering
            \scriptsize
            \setlength{\tabcolsep}{3pt}
            \resizebox{\linewidth}{!}{
            \begin{tabular}{l|l|cccccc|c}
            \toprule
            \multicolumn{2}{c|}{Model}& \parbox[t]{2mm}{\rotatebox[origin=c]{90}{ MiniGPT-4}}  & \parbox[t]{2mm}{\rotatebox[origin=c]{90}{ InstructBLIP}}  & \parbox[t]{2mm}{\rotatebox[origin=c]{90}{ BLIP2}}  & \parbox[t]{2mm}{\rotatebox[origin=c]{90}{ VPGTrans}}  & \parbox[t]{2mm}{\rotatebox[origin=c]{90}{ BLIVA}}  & \parbox[t]{2mm}{\rotatebox[origin=c]{90}{ VisualGLM}} &  Avg.\\\midrule
            \multicolumn{9}{c}{\bf ImageNette} \\ \midrule
            \multirow{5}{*}{VP~\cite{bahng2022exploring}} & 1\% & 77.58 & 64.08 & 93.50 & 77.73 & 73.53 & 72.25 & 76.45 \\
 & 5\% & 81.91 & 64.28 & 91.80 & 81.40 & 80.48 & 73.45 & 78.89 \\
 & 10\% & 78.22 & 65.17 & 95.13 & 77.12 & 67.52 & 72.43 & 75.93 \\
 & 25\% & 82.42 & 60.25 & 93.12 & 79.29 & 71.41 & 76.10 & 77.10 \\
 & 50\% & 82.01 & 62.14 & 92.64 & 76.66 & 77.83 & 76.94 & 78.04 \\\midrule
\multirow{5}{*}{EVP~\cite{wu2022unleashing}} & 1\% & 93.01 & 62.00 & 94.96 & 74.07 & 83.38 & 80.57 & 81.33 \\
 & 5\% & 97.61 & 62.80 & 89.10 & 77.10 & 79.88 & 80.25 & 81.12 \\
 & 10\% & 98.00 & 71.87 & 90.37 & 72.97 & 76.05 & 75.85 & 80.85 \\
 & 25\% & 97.44 & 76.08 & 89.15 & 63.99 & 84.58 & 74.24 & 80.91 \\
 & 50\% & 97.40 & 62.06 & 85.12 & 74.70 & 82.52 & 86.57 & 81.39 \\\midrule
\multirow{5}{*}{TVP (ours)} & 1\% & 96.13 & 72.08 & 89.30 & 79.06 & 90.78 & 70.45 & 82.97 \\
 & 5\% & 97.61 & 62.70 & 91.80 & 76.87 & 83.46 & 91.95 & 84.07 \\
 & 10\% & 97.20 & 72.25 & 94.70 & 83.21 & 82.93 & 74.68 & 84.16 \\
 & 25\% & 97.63 & 72.48 & 89.86 & 85.96 & 80.46 & 73.71 & 83.35 \\
 & 50\% & 98.09 & 73.99 & 90.29 & 84.66 & 86.09 & 75.03 & 84.69\\\midrule
        \multicolumn{9}{c}{\bf CLEVR} \\ \midrule
        \multirow{5}{*}{VP~\cite{bahng2022exploring}} & 1\% & 33.17 & 31.57 & 12.73 & 12.73 & 32.87 & 12.87 & 22.66 \\
 & 5\% & 38.37 & 27.57 & 12.83 & 21.63 & 19.23 & 12.57 & 22.03 \\
 & 10\% & 40.13 & 36.00 & 23.33 & 10.97 & 28.67 & 12.63 & 25.29 \\
 & 25\% & 39.81 & 26.44 & 12.95 & 12.50 & 21.84 & 13.77 & 21.22 \\
 & 50\% & 39.30 & 28.80 & 12.77 & 12.00 & 28.63 & 12.83 & 22.39 \\\midrule
\multirow{5}{*}{EVP~\cite{wu2022unleashing}} & 1\% & 47.03 & 35.53 & 15.47 & 11.37 & 31.80 & 13.03 & 25.71 \\
 & 5\% & 25.07 & 35.97 & 27.93 & 9.77 & 34.60 & 14.30 & 24.61 \\
 & 10\% & 49.80 & 34.57 & 16.43 & 15.60 & 32.40 & 13.10 & 26.98 \\
 & 25\% & 44.70 & 32.43 & 20.67 & 16.40 & 35.57 & 12.90 & 27.11 \\
 & 50\% & 53.70 & 42.60 & 20.10 & 7.47 & 33.87 & 12.90 & 28.44 \\\midrule
\multirow{5}{*}{TVP (ours)} & 1\% & 45.60 & 37.70 & 15.13 & 25.50 & 34.70 & 12.77 & 28.57 \\
 & 5\% & 43.10 & 17.50 & 26.60 & 15.03 & 41.80 & 23.17 & 27.87 \\
 & 10\% & 48.77 & 45.93 & 14.53 & 20.60 & 37.67 & 13.40 & 30.15 \\
 & 25\% & 47.13 & 42.77 & 19.80 & 21.13 & 34.53 & 12.87 & 29.71 \\
 & 50\% & 47.37 & 42.37 & 24.43 & 15.70 & 35.27 & 13.30 & 29.74  \\\midrule
        \multicolumn{9}{c}{\bf POPE} \\ \midrule
        \multirow{5}{*}{VP~\cite{bahng2022exploring}} & 1\% & 54.67 & 68.13 & 49.87 & 58.07 & 72.53 & 69.67 & 62.16 \\
 & 5\% & 51.53 & 66.87 & 50.00 & 62.87 & 73.73 & 71.53 & 62.76 \\
 & 10\% & 52.73 & 70.67 & 50.00 & 59.07 & 73.00 & 69.13 & 62.43 \\
 & 25\% & 53.93 & 70.73 & 49.87 & 59.27 & 73.27 & 71.60 & 63.11 \\
 & 50\% & 51.00 & 71.27 & 50.00 & 63.07 & 73.93 & 69.33 & 63.10 \\\midrule
\multirow{5}{*}{EVP~\cite{wu2022unleashing}} & 1\% & 51.60 & 74.00 & 50.00 & 58.67 & 72.53 & 68.13 & 62.49 \\
 & 5\% & 60.73 & 67.60 & 50.04 & 61.24 & 74.77 & 69.47 & 63.97 \\
 & 10\% & 64.73 & 65.33 & 50.00 & 59.67 & 71.67 & 70.47 & 63.65 \\
 & 25\% & 58.73 & 73.93 & 50.00 & 60.67 & 74.20 & 69.47 & 64.50 \\
 & 50\% & 61.36 & 72.13 & 49.95 & 61.27 & 72.69 & 69.98 & 64.56 \\\midrule
\multirow{5}{*}{TVP (ours)} & 1\% & 61.00 & 71.47 & 50.13 & 59.00 & 75.00 & 69.80 & 64.40 \\
 & 5\% & 62.73 & 69.47 & 50.00 & 65.20 & 72.47 & 70.53 & 65.07 \\
 & 10\% & 59.87 & 75.47 & 49.87 & 65.13 & 74.27 & 70.67 & 65.88 \\
 & 25\% & 71.13 & 69.93 & 49.60 & 60.60 & 72.67 & 70.40 & 65.72 \\
 & 50\% & 70.47 & 68.67 & 49.87 & 61.00 & 75.47 & 70.33 & 65.97  \\\bottomrule
        \end{tabular}}
    \end{subtable}
    \caption{Detailed results for the analysis on the impact from different training data scales in~\cref{fig:data-scale}.}
\label{tab:detailed_datascale}
\end{table*}



\end{document}